%% file: arxiv.tex
\documentclass[10pt,twocolumn,letterpaper]{article}

\usepackage[pagenumbers]{cvpr} 
\usepackage[accsupp]{axessibility}
\usepackage{graphicx}
\usepackage{amsmath}
\usepackage{amssymb}
\usepackage{booktabs}
\usepackage{cite}
\usepackage{amsmath,amssymb,amsfonts}
\usepackage{algorithmic}
\usepackage{graphicx}
\usepackage{textcomp}
\usepackage{tabularx}
\usepackage{url}
\usepackage{times}
\usepackage{epsfig}
\usepackage{graphicx}
\usepackage{amsmath}
\usepackage{amssymb}
\usepackage{stfloats}
\usepackage{multirow}
\usepackage{colortbl}
\usepackage{float}
\usepackage{color}
\usepackage[section]{placeins}
\definecolor{mygray}{RGB}{193,203,215}
\usepackage{bigstrut}
\usepackage{amsmath}

\input{preamble}

\definecolor{cvprblue}{rgb}{0.21,0.49,0.74}
\usepackage[pagebackref,breaklinks,colorlinks,citecolor=cvprblue]{hyperref}

\def\paperID{16} 
\def\confName{CVPR}
\def\confYear{2024}

\title{Generalized Single-Image-Based Morphing Attack Detection Using Deep Representations from Vision Transformer}

\author{Haoyu Zhang$^1$\thanks{This work was supported by the European Union’s Horizon 2020 Research and Innovation Program under Grant 883356.} \and Raghavendra Ramachandra$^1$ \and Kiran Raja$^1$ \and Christoph Busch$^{1,2}$  \\
$^1$ Norwegian University of Science and Technology, Norway\\  
	$^2$ Darmstadt University of Applied Sciences, Germany\\
	\{\tt\small haoyu.zhang, raghavendra.ramachandra, kiran.raja, christoph.busch\}@ntnu.no\\
	\tt\small christoph.busch@h-da.de\\
}
\begin{document}
\maketitle
\begin{abstract}
Face morphing attacks have posed severe threats to Face Recognition Systems (FRS), which are operated in border control and passport issuance use cases. Correspondingly, morphing attack detection algorithms (MAD) are needed to defend against such attacks. MAD approaches must be robust enough to handle unknown attacks in an open-set scenario where attacks can originate from various morphing generation algorithms, post-processing and the diversity of printers/scanners. The problem of generalization is further pronounced when the detection has to be made on a single suspected image. In this paper, we propose a generalized single-image-based MAD (S-MAD) algorithm by learning the encoding from Vision Transformer (ViT) architecture. Compared to CNN-based architectures, ViT model has the advantage on integrating local and global information and hence can be suitable to detect the morphing traces widely distributed among the face region. Extensive experiments are carried out on face morphing datasets generated using publicly available FRGC face datasets. Several state-of-the-art (SOTA) MAD algorithms, including representative ones that have been publicly evaluated, have been selected and benchmarked with our ViT-based approach. Obtained results demonstrate the improved detection performance of the proposed S-MAD method on inter-dataset testing (when different data is used for training and testing) and comparable performance on intra-dataset testing (when the same data is used for training and testing) experimental protocol.
\end{abstract}
\section{Introduction}
Face recognition systems (FRS) have been widely deployed in various security applications, such as passport issuance and automated border control (ABC)\cite{jain2011handbook}. However, with the development of image manipulation techniques, FRS are becoming vulnerable to different kinds of attacks that may lead to security lapses \cite{scherhag2017biometric} \cite{9380153}. Morphing attack is one type of the attacks that targets to subvert FRS by combining biometric samples from 2 or more individuals into a single morphed image. Morphing attacks have been illustrated as an evolving threat to the FRS \cite{ferrara2014magic}. Morphing attack detection algorithms (MAD) have been therefore proposed to detect these attacks to improve the security of FRS.

\begin{figure*}[htp]
    \centering
   \includegraphics[width=1\linewidth]{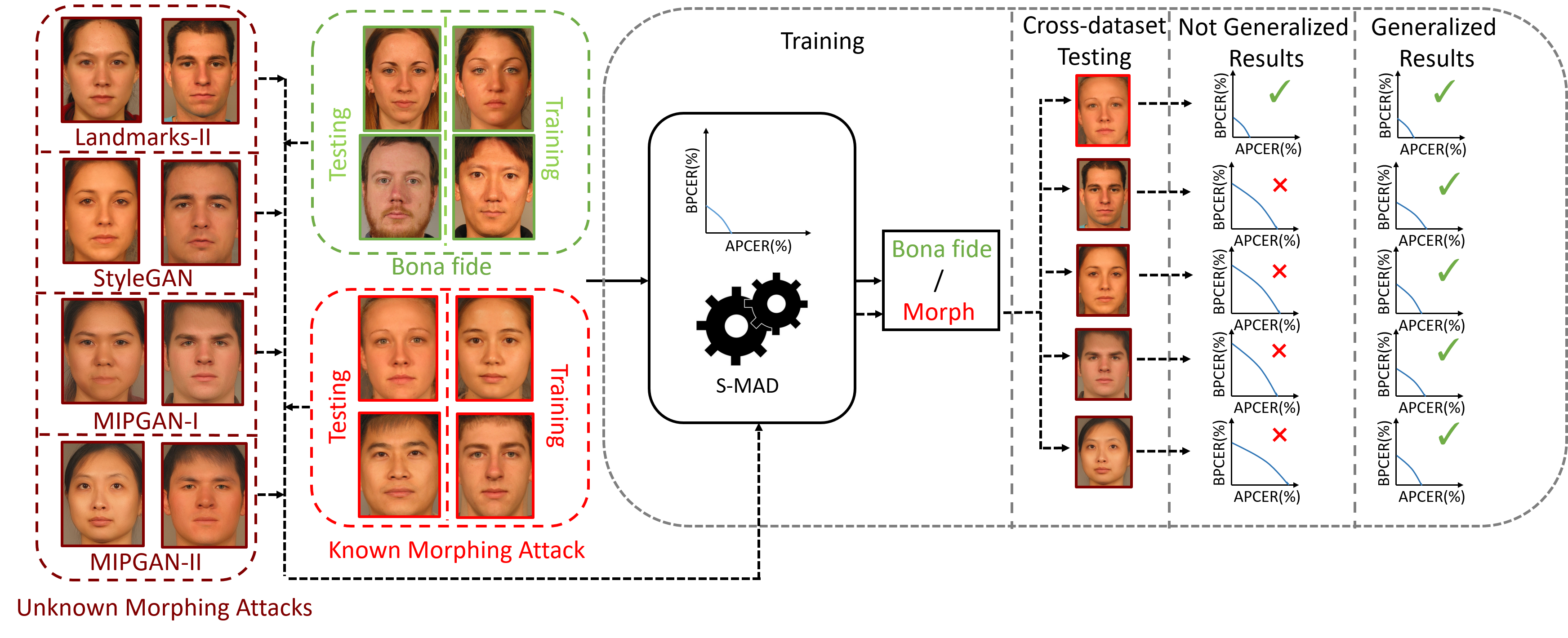}
   \caption{Hypothesised illustration of S-MAD as open-set problem: A model trained on known morphing attacks may fail at unknown morphing attacks.}
\label{fig:intro}
\end{figure*}
Single-image-based morphing attack detection (S-MAD) aims to detect the face morphing attack based on a single image presented to the algorithm. The most common application scenario of S-MAD is validating the face photos submitted in passport or visa applications (physically/through online services) \cite{9380153}. Another possible used case for S-MAD is the validation of an existing face image database, to validate that no morphed images are contained. Hence, the S-MAD algorithm should well generalize for different types of face images and anticipated image processing, such as digital, print-scanned and print-scanned-compression. In addition, there are various types of morphing algorithms that generate morphed face images with different characteristics, such as realistic texture and high face structure similarity. While many previous works have developed MAD approaches that can detect attacks efficiently for known kinds of morphing attacks, the performance tends to degrade when testing involves data stemming from different morphing methods and which were unseen during training. \cref{fig:intro} illustrates an example of such a scenario when the S-MAD algorithm trained on the known attack (i.e., known morphing generation type) can easily miss detecting an attack from the unknown generation type \cite{9246583}. Given the envisioned application scenario, it is crucial to improve the generalizability of S-MAD algorithm and to evaluate the detection performance in an open-set scenario by cross-dataset testing.

The existing S-MAD approaches are based on texture features \cite{scherhag2018morph}, residual noise features, hybrid features, and deep learning features \cite{9380153} \cite{ferrara2019face} \cite{ramachandra2020detecting} \cite{seibold2019style}. With the achievement of deep convolutional neural networks (CNNs) in the field of image recognition, many researchers have applied pretrained CNNs and transfer learning to solve the S-MAD problems as binary classification problems \cite{raja2017transferable} \cite{ferrara2019face}. Although it has been shown that CNN-based methods may achieve better performance than S-MAD methods based on hand-crafted features, the generalizability of these approaches to print-scan images tends to be limited \cite{9246583}. 

Recently, Vision Transformer (ViT) \cite{dosovitskiy2020image} has become popular in computer vision and has achieved impressive results on existing image recognition challenges.  Transformer models \cite{vaswani2017attention} apply the concepts of natural language processing directly to images where an image is split into small patches and then projected as a sequence of linear embeddings, which further are treated as the input to a Transformer model. By applying the self-attention mechanism and without introducing strong image-specific inductive biases as CNNs, ViT has shown the capability to integrate information globally from low layers and has achieved state-of-the-art (SOTA) performance in different tasks with large-scale training data. Consequently, many works have been investigating the possibility of applying ViT to other tasks. In the case of MAD, the traces of morphing are widely distributed among the face region, and hence the algorithms should have a large receptive field and the capacity of integrating local and global information to be robust and generalized. Hence, We assert that the advantages of ViTs can improve S-MAD and investigate further if they improve the generalizability of the developed S-MAD algorithm.

\textbf{Our Contributions}: 1) We propose an S-MAD algorithm based on the deep representation from a pretrained vanilla ViT against other works using CNNs.
2) We investigate the applicability of the pure self-attention-based model in S-MAD tasks by conducting comprehensive cross-dataset testing with various morph generation types and different dataset types (digital/print-scan/print-scan compression). The generalizability and detection performance of the proposed approach is quantitatively evaluated and reported 3) We benchmark the proposed method together with other state-of-the-art S-MAD algorithms based on the ensemble of hand-crafted features \cite{EnsembleFeatures_2020}, hybrid scale-space colour texture features \cite{RagISBA2019} (reported in the testing report from National Institute of Standards and Technology \cite{ngan2021face}), deep CNN features \cite{raja2017transferable}, steerable features \cite{ramachandra2019detecting}, Multi-modality approach (tested in Bologna Online Evaluation Platform \cite{raja2020morphing}\footnote{\url{https://biolab.csr.unibo.it/fvcongoing/UI/Form/BOEP.aspx}}, residual AutoEncoder \cite{raja2022towards}, and Multi-level Deep Features \cite{venkatesh2022multi} respectively. The analysis result indicates an improved generalizability on digital inputs.


\section{Proposed Method}

\begin{figure*}[htp]
    \centering
   \includegraphics[width=1\linewidth]{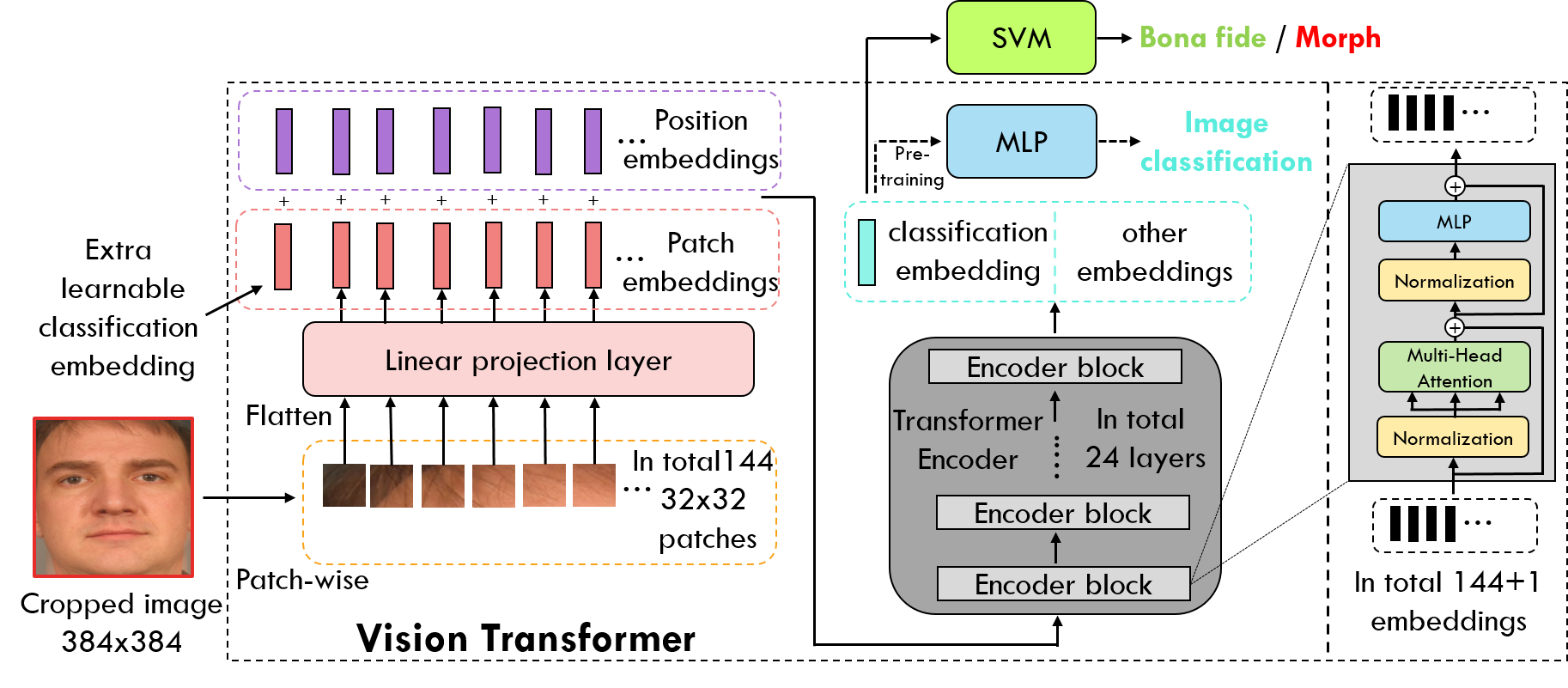}
   \caption{Overview of our proposed method using pretrained Vision Transformer model.}
\label{fig:overview}
\end{figure*}
An overview of our proposed S-MAD method is described in \cref{fig:overview}. We first crop the face region using MTCNN \cite{zhang2016joint} to detect face regions and then resize the cropped face image into 384 x 384 pixels to fit the input of ViT model. Then, the input image is split into small patches $x_p$ with the size of 32 x 32 pixels and then flattened and projected as patch embeddings through a learned linear projection layer with one layer of fully connected blocks for each embedding.
Then an extra learnable classification embedding $x_{class}$ is attached to the other patch embeddings as the learned image representation for further classification tasks. Similar to the design of the vanilla ViT, 1-D positional encoding is applied to generate position embeddings $E_{pos}$ with the same length of the patch embeddings using sinusoidal functions. Each position embedding is added to the corresponding patch embedding hence the positional information can be encoded. Then the processed input $z_0$ can be noted as:
\begin{equation}
    z_0 = [x_{class}; x_{p}^{1}E;x_{p}^{2}E;...;x_{p}^{N}E]+E_{pos}
\end{equation}
where $N=144$ is the number of patches and $E$ indicates the linear projection process.
After processing the image into a sequence of embeddings, they are fed forward through the transformer encoder stacked with 24 layers of encoder blocks. Each encoder block includes a multi-head self-attention layer and a Multilayer Perceptron (MLP) block.
\begin{align}
    z_{l}' = MSA(LN(z_{l-1}))+z_{l-1}, \;l=1,...,L.
\end{align}
\begin{align}
    z_l = MLP(LN(z_{l}'))+z_{l}',  \;l=1,...,L.
\end{align}
The multi-head self-attention layer extends the key-query-value triplet into 16 sub-triplets and executes the computation of the self-attention mechanism in parallel, hence the model can learn to extract features from multiple different aspects. 

During Pretraining of the ViT model, the classification token is linked to an extra MLP with a dimension of 4096 and then a softmax classifier for image classification. The model is pretrained on ImageNet21k \cite{ridnik2021imagenet} and ImageNet2012 dataset \cite{russakovsky2015imagenet} with 1000 classes. To avoid duplicated training processes and achieve sustainability, we use the settings of hyper-parameters inspired by the original ViT paper \cite{dosovitskiy2020image}. As for model selection, we selected the ViT-L model with the large parameter size for higher capacity generalizability and large patch size to extract more local information. For the S-MAD task, we use the pretrained model to extract the classification tokens with the dimension of 1024 on our face morphing dataset. The extracted classification tokens will be considered as general deep representations and then we train a linear SVM classifier to solve the S-MAD problem as a binary classification task. The SVM classifier is chosen over training a deep-learning-based binary classifier due to its efficiency and robustness in preventing overfitting for small-to-medium size datasets.


\section{Dataset}

In order to conduct the cross-dataset testing comprehensively and simulate the operational use case, we use a database generated by various morphing algorithms and in different image processing methods (digital, print-scanned, print-scanned-compression). To simulate the passport use cases with face photos, our database is constructed based on selected morphed images from FRGC-V2 dataset \cite{FRGC_DB} with high image quality and well-controlled capturing conditions (e.g., pose variations). 140 unique subjects, including 93 male subjects and 47 female subjects are selected. For each subject, 7-24 mated samples with similar capturing conditions (image resolution, neutral expression, pose, illumination, etc.) are chosen. In total, 1270 bona fide samples are included.

As for the morphing algorithms, we selected the following five representative morphing algorithms including, two landmark-based algorithms Landmark-I \cite{raghavendra2017face} and Landmark-II \cite{UBO_Morphing_Tool}, and 3 GAN-based algorithms StyleGAN-IWBF\cite{MorphStyleGAN2020}, MIPGAN-I and MIPGAN-II \cite{zhang2021mipgan} to establish a diversity of unknown attacks. The samples are pre-processed to meet the ICAO 9303 requirements \cite{ICAO_9303_p9_2021}. Pairs of parent images for the morphing process are selected following guidelines suggested in \cite{raghavendra2017face} \cite{scherhag2017biometric} (e.g., isolating between different genders, pairing based on similarity score of an FRS model), as the attacker may spend as much as an effort to generate the morphing attacks in real cases. As our target is to train the model to learn patterns generated from morphing instead of general patterns from GANs, reconstructed bona fide images are applied to the datasets with GAN-based morphing algorithms. In this way, we can reduce the bias between bona fide and morph samples and can make the trained classifier generalize to other types of attacks that are not generated by the same GAN model.

To evaluate the generalizability of S-MAD algorithms on different types of images, 3 types are included in our database:
\begin{itemize}
    \item \textbf{Digital}: Morph images are obtained from the morphing algorithms given digital parent images as input.
    \item \textbf{Print-scan}: Both generated morphs and bona fide images are printed using DNP-DS820 dye-sublimation photo printer and then re-digitized using the Canon office scanner with 300 dpi as suggested in ICAO 9303 requirements \cite{ICAO_9303_p9_2021}. This is to simulate the process of a passport application.
    \item \textbf{Print-scan with compression}: Print-scanned images (morphs and bona fide) are compressed into less or equal to 15 KBs to simulate the images stored in the e-passport.
\end{itemize}
Overall, each dataset has $2500$ morphed images and $1270$ bona fide images. Given the 5 included morphing algorithms and 3 image processing types, in total 15 datasets are used in the database for further cross-dataset testing on S-MAD algorithms.

\section{Experiments and Results}
\label{sec:exp}
To evaluate the generalizability and robustness of our approach, we apply cross-dataset testing on different morphing algorithms within each dataset of different image processing types and benchmark it with the other selected SOTAs:
\begin{itemize}
    \item Ensemble Features \cite{EnsembleFeatures_2020} uses ensembled features including LBP, HoG, and BSIF. The algorithm has been evaluated by public testing and included in NIST report \cite{ngan2021face}
    \item Hybrid Features \cite{RagISBA2019} uses scaled colour space and trains independent classifiers based on the extracted LBP features.
    \item Deep Features \cite{raja2017transferable} use pretrained VGG and AlexNet to extract transferable features and apply feature-level fusion for further classification. 
    \item Steerable Features \cite{ramachandra2019detecting} extracts steerable pyramids from illuminance components and trains classifiers based on high-frequency components.
    \item Multi-Modality \cite{raghavendra2022multimodality} crops the face image into different regions and extracts BSIF and LBP features. Independent classifiers are trained and score-level fusion is applied to output the final classification result. The algorithm has been evaluated in the Bologna Online Evaluation Platform \cite{raja2020morphing} \footnote{\url{https://biolab.csr.unibo.it/FvcOnGoing/UI/Form/AlgResult.aspx?algId=8422}}. 
    \item Residual AutoEncoder \cite{raja2022towards} is a deep learning approach that consists of a skip-connected AutoEncoder and a ResNet18 Classifier. Guided by the designed loss functions, the model is trained to extract learnable residuals which can be used for further classification by the ResNet.
    \item Multi-level Deep Features \cite{venkatesh2022multi} applies multi-level fusion on features extracted from AlexNet and ResNet50.
\end{itemize}
The selected baselines cover approaches based on hand-crafted features, deep-learning-based transferable features, different fusion strategies, and trained deep-learning models.

More specifically, for each dataset generated with a specific morphing algorithm, we train the S-MAD algorithm on it and test with the datasets (generated by different morphing algorithms). This shows how the detection algorithms can generalize and to which extent they are robust with respect to unknown attacks. The performance of testing across different image processing types is not included as considering a model trained on print-scan data is often not used to detect attacks from digital data rather an ensemble is used. Instead, we report the performance of cross-dataset testing for the same image processing types (e.g., digital versus digital) to evaluate the generalizability of MAD algorithms.

To report the performance of each test, we employ standardized metrics such as Bona fide Presentation Classification Error Rate (BPCER) and Morphing Attack Classification Error Rate (MACER) following ISO/IEC CD 20059.2 \cite{ISO-IEC-20059-2023} and measure the detection error trade-off by reporting BPCER@MACER=5$\%$ and BPCER@MACER=10$\%$. To simplify and scalarize the results, Detection equal error rate (D-EER) is also reported. The lower D-EER numbers indicate better detection performances.

\begin{figure*}[h!]
    \centering
   \includegraphics[width=0.95\linewidth]{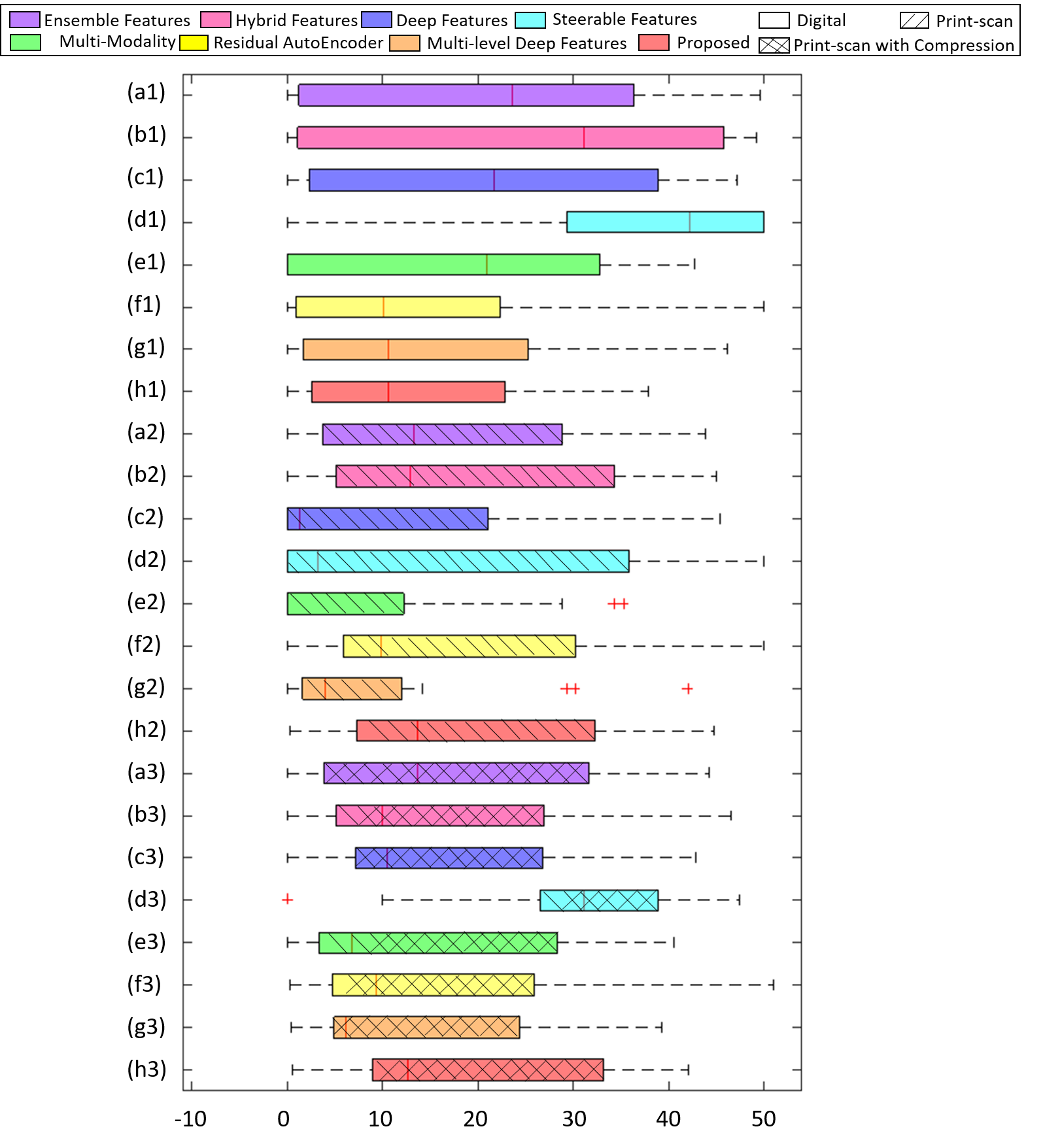}
   \caption{Boxplot of the statistical analysis on D-EER computed for all cross-dataset testing results on FRGC morph database. (a) Ensemble Features (b) Hybrid Features  (c) Deep Features (d) Steerable Features(e) Multi-Modality (f) Residual AutoEncoder (g) Proposed Method. (1): Digital (2): Print-scan (3): Print-scan Compression}
\label{fig:boxplot}
\end{figure*}

For the evaluation protocol, we evaluate both intra-dataset testing and inter-dataset testing but without crossing image types (digital, print-scanned, and print-scanned and compressed). Detailed quantitative analysis is included in the supplementary material. To measure the overall generalizability of the MAD algorithms and establish the significance of the obtained results, we propose to conduct statistical analysis on the D-EER of the cross-dataset testing cases within each type of image, and also visualize this analysis as a boxplot. 
From the quantitative analysis in \cref{tab:statistical}, it is shown that our approach has the lowest mean and standard deviation of D-EER for digital images. The mean value of D-EERs from the proposed method has decreased 1.15$\%$ compared to the best among the baselines. As visualized in \cref{fig:boxplot}, a similar observation can also be noticed by the similarly low median value as Residual AutoEncoder and Multi-level Deep Features. However, the range of error rate from our approach during testing is more narrowed, which indicates better robustness. In print-scan and print-scan with compression cases, a degradation of the detection performance of our algorithm can be noticed compared to the digital case. We reason this by 1) the Vision Transformer model is pretrained only with digital images and hence the extracted representation is less effective when transferred to another image processing type 2) compared to the digital images, print-scan and print-scan compression images are in a much lower resolution and can provide less information for the Vision Transformer model (which takes the input size of 384 x 384). For print-scan inputs, the Multi-modality approach and multi-level Deep Features approach achieved the best performances. As for further compressed print-scanned images, the multi-modality approach and Residual AutoEncoder approach are similar. 

\begin{table*}[hbtp]
\centering
 \caption{Statistical analysis on the D-EER($\%$) computed for all cross-dataset testing results on FRGC morphing dataset.}
\resizebox{0.75\linewidth}{!}{
\begin{tabular}{|c|p{1.1cm}<{\centering}|p{1cm}<{\centering}|p{1cm}<{\centering}|p{1cm}<{\centering}|p{1.4cm}<{\centering}|p{1.4cm}<{\centering}|}
\hline
\textbf{S-MAD} & \multicolumn{2}{c|}{\textbf{Digital}}    & \multicolumn{2}{c|}{\textbf{Print-scan}} & \multicolumn{2}{c|}{\textbf{P.S. with Compression}} \\ 
\cline{2-7} \textbf{Algorithms} & $\mu$ & $\sigma$ & $\mu$ & $\sigma$ & $\mu$    & $\sigma$   \\ \hline
Ensemble & \multicolumn{1}{c|}{21.03} & 17.55 & \multicolumn{1}{c|}{16.73} & 14.90 & \multicolumn{1}{c|}{17.24} & 15.78 \\ 
Features \cite{EnsembleFeatures_2020} &  & &  &  &  &  \\ \hline
Hybrid   & \multicolumn{1}{c|}{26.11} & 20.26 & \multicolumn{1}{c|}{19.07} & 16.04 & \multicolumn{1}{c|}{16.28} & 14.23 \\ 
Features \cite{RagISBA2019}  &  &  &  &  &  &  \\ \hline
Deep    & \multicolumn{1}{c|}{21.16} & 17.50 & \multicolumn{1}{c|}{11.38} & 16.23 & \multicolumn{1}{c|}{16.77} & 13.19 \\ 
 Features \cite{raja2017transferable}  &  &  &  &  &  &  \\ \hline
 Steerable    & \multicolumn{1}{c|}{35.97} & 16.57 & \multicolumn{1}{c|}{15.72} & 18.44 & \multicolumn{1}{c|}{31.49} & 11.20 \\ 
 Features\cite{ramachandra2019detecting}  &  &  &  &  &  &  \\ \hline
Multi-   & \multicolumn{1}{c|}{18.05} & 15.51 & \multicolumn{1}{c|}{\textbf{7.63}} & \textbf{12.27} & \multicolumn{1}{c|}{13.57} & 13.72 \\ 
Modality \cite{raghavendra2022multimodality}  &  &  &  &  &  &  \\ \hline
Residual   & \multicolumn{1}{c|}{15.95} & 17.38 & \multicolumn{1}{c|}{16.01} & 15.25 & \multicolumn{1}{c|}{14.92} & 13.97 \\ 
AutoEncoder \cite{raja2022towards}  &  &  &  &  &  &  \\ \hline
Multi-level & \multicolumn{1}{c|}{14.78} & 13.90 & \multicolumn{1}{c|}{9.51} & 12.44& \multicolumn{1}{c|}{\textbf{13.38}} & \textbf{12.36}  \\ 
Deep Feature \cite{venkatesh2022multi}  &  &  &  &  &  &  \\ \hline
\textbf{Proposed Method}   & \multicolumn{1}{c|}{\textbf{13.63}} & \textbf{11.61} & \multicolumn{1}{c|}{18.33} & 14.32 & \multicolumn{1}{c|}{19.09} & 13.61 \\ \hline
\end{tabular}
}
\label{tab:statistical}%
\end{table*}
It is also shown that different algorithms perform inconsistently for the same testing cases. Ensemble Features, Hybrid Features, Deep Features, and Steerable Features are not generalizing well for most of the cross-testing cases, even for inter-testing among GAN-based morphed images. Our approach has shown considerable generalization when the test is crossing between landmark-based and GAN-based morphs in digital cases. The Residual AutoEncoder approach has also shown similar performance, while some extreme cases can be noticed for example in when the model is trained by MIPGAN-I dataset and tested on Landmark-II dataset. This might be caused by the randomness during the training process of the network. The multi-Modal approach in general has impressive performances on Print-scan and print-scan compression images.
Compared with Deep Features approach with transferable CNN features, it is shown that the ViT-based features can achieve improvement in the generalizability of MAD for digital images. Meanwhile, the multi-level fusion of CNN features has shown an overall improvement in the Deep Features approach. It also achieves comparable results with ViT-based features in the digital case with considerable generalizability for print-scan and print-scan compression images.

\begin{figure}[h]
    \centering
   \includegraphics[width=0.95\linewidth]{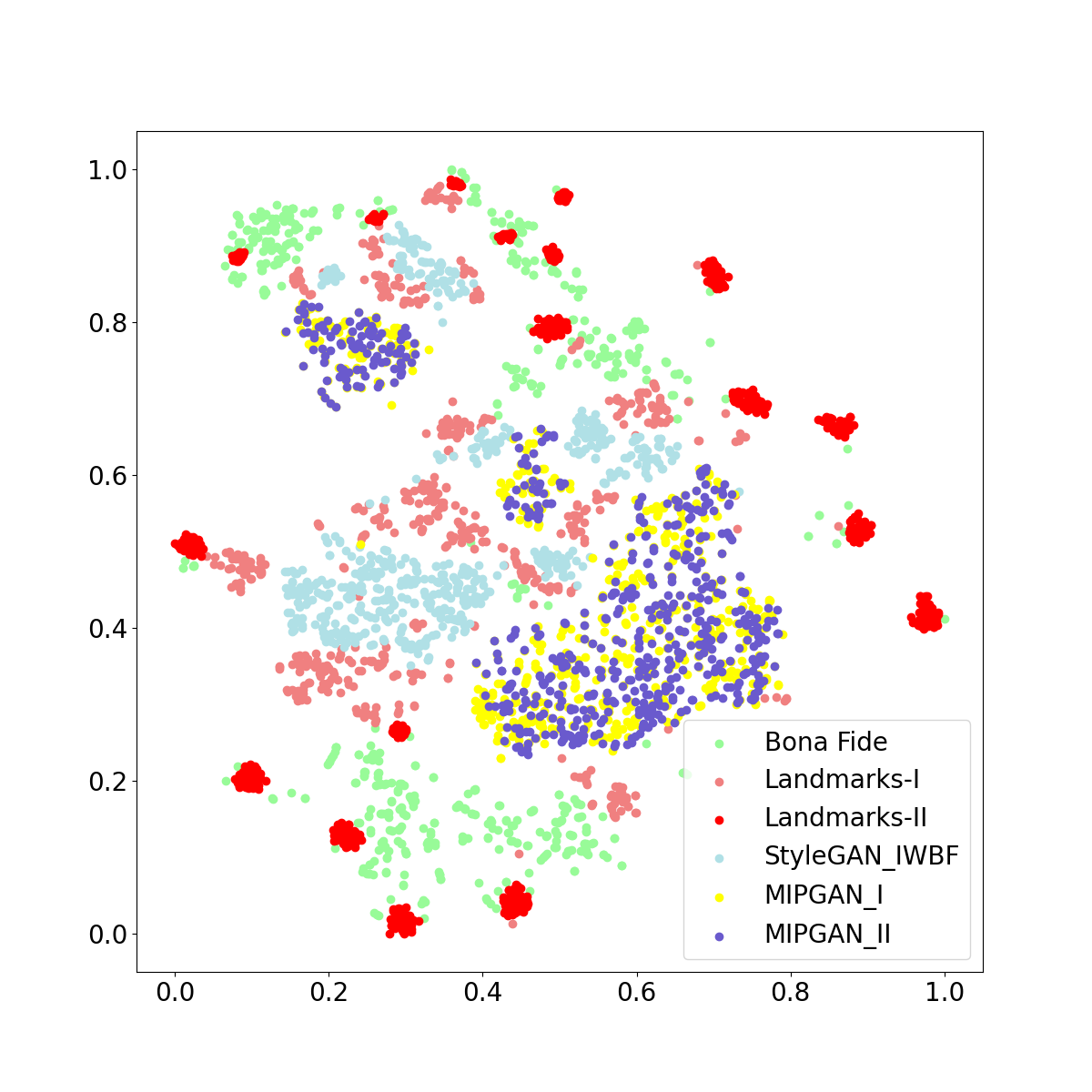}
   \caption{T-SNE plot of the feature space used in proposed method with digital images}
\label{fig:tsne_d}
\end{figure}

\begin{figure}[h]
    \centering
   \includegraphics[width=0.95\linewidth]{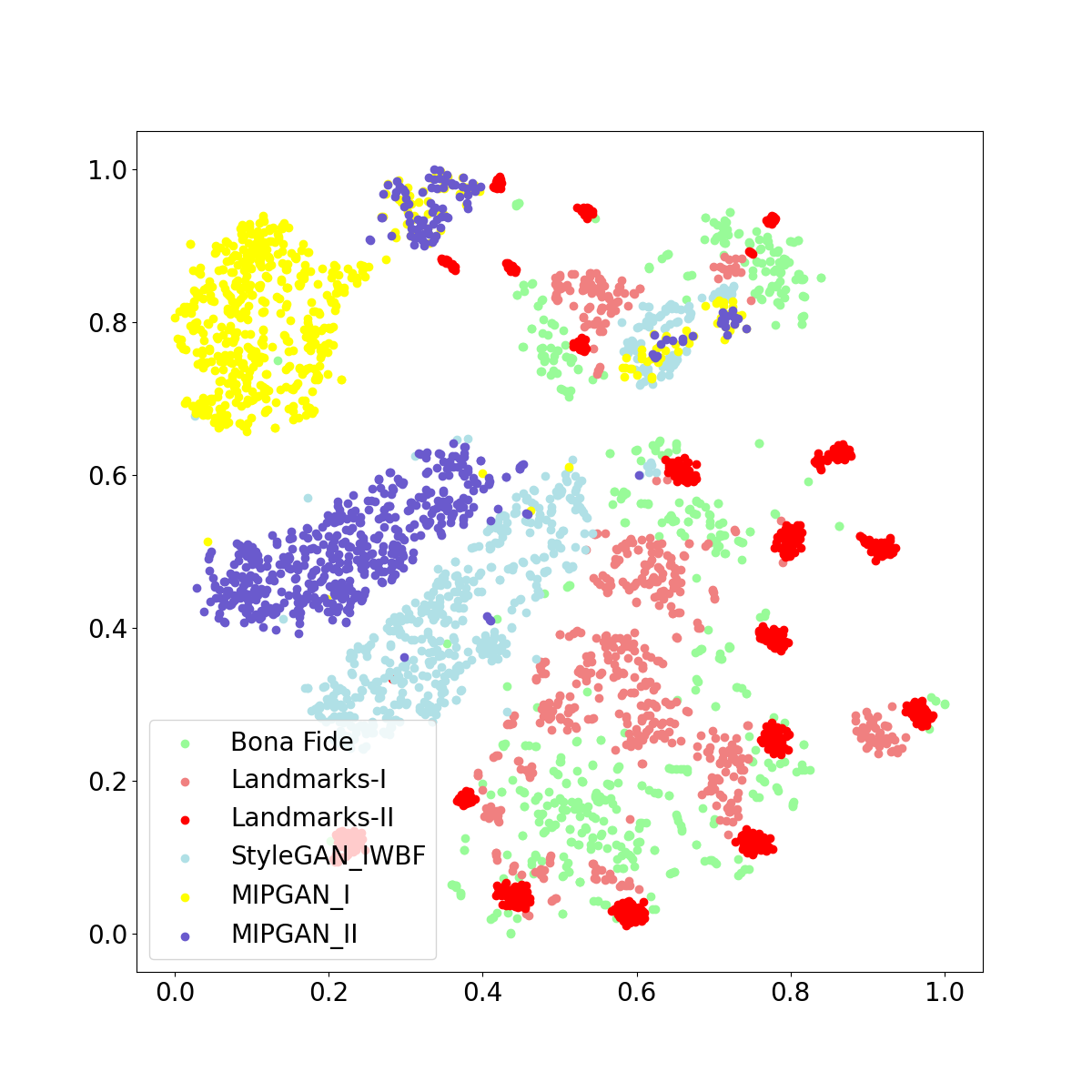}
   \caption{T-SNE plot of the feature space used in proposed method with print-scanned images}
\label{fig:tsne_ps}
\end{figure}

\begin{figure}[h]
    \centering
   \includegraphics[width=0.95\linewidth]{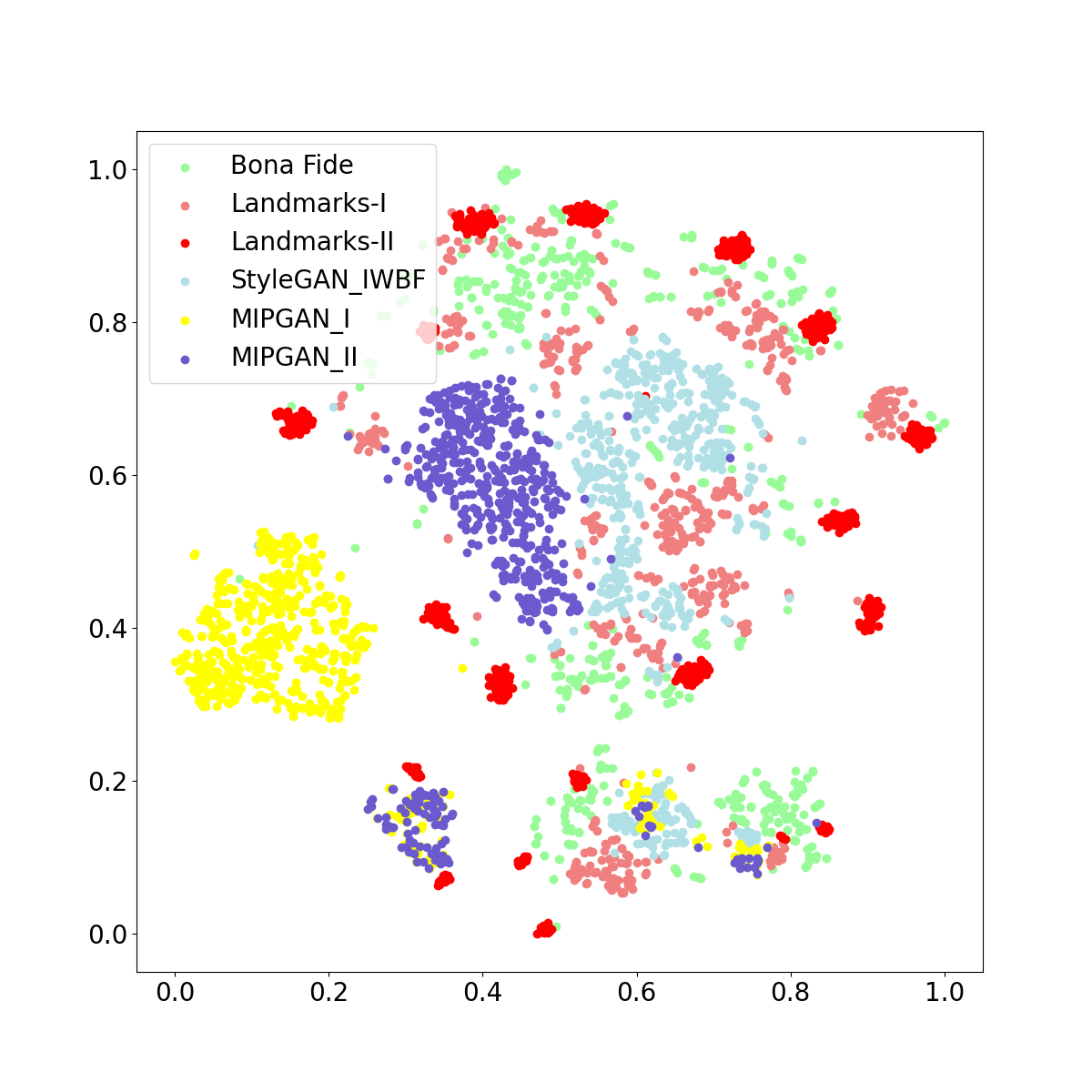}
   \caption{T-SNE plot of the feature space used in proposed method with print-scanned and compressed images}
\label{fig:tsne_psc}
\end{figure}

Additionally, the interpretation of the proposed method and obtained results is studied. As shown in \cref{fig:tsne_d} - \cref{fig:tsne_psc}, T-SNE \cite{van2008visualizing} plot is used to visualize the feature space of the proposed method with data using different processing processes. 

Similarly as what we've observed in the cross-dataset testing results, features from morphs generated by StyleGAN-IWBF, MIPGAN-I and MIPGAN-II can be well-separated between the features from bona fide images, which indicates good generalization on cross-dataset testing. Features of Landmark-I morphs are shown to be less seperable than GAN-based morphs. The overlap between features of morphed samples generated by Landmark-II method and features of bona fide samples also follows detection accuracy where this method is the most challenging one to classify and generalize. This shows that the post-processing on morphs can effectively make the generated attacks stronger. Meanwhile, when the processing type is changed from digital (\cref{fig:tsne_d}) to print-scan (\cref{fig:tsne_ps}) and then to print-scan compression (\cref{fig:tsne_psc}), it becomes more difficult to find sharable boundaries for classifying different types of morphs between bona fide samples.

\section{Limitations}

In this work we have applied a specific pretrained ViT model in order to be sustainable on computational powers, however, the influence of different hyper-parameters in the ViT model on the final performance of S-MAD tasks is still worth to be studied. Meanwhile, for the cross-dataset testing, we only conducted experiments with leave-one-out training. It is also interesting to evaluate S-MAD trained on a dataset mixed with multiple morphing algorithms and study on the learning capacity. By using the pretrained ViT model, our model has shown an improvement in generalizability for digital images, while it can be noticed that for intra-dataset testing the detection accuracies are overall less or equal for the other algorithms. Also as shown in our evaluation, the different algorithm performs inconsistently. Hence, it is reasonable to further explore fusion strategies or combine them with the multi-modality approach.

\section{Conclusion}

In this work, we proposed an S-MAD algorithm based on pretrained Vision Transformer model instead of existing deep-learning-based methods using CNNs. Motivated by the real application scenario of open-set testing, we use a morph dataset with three different image processing types and five different representative morphing algorithms, including both GAN-based and landmark-based algorithms for the cross-dataset testing. The proposed method is benchmarked against two selected SOTA algorithms. Based on the statistical analysis of the obtained results, it can be concluded that the proposed method based on the pure self-attention model can achieve notable improvement in the generalizability of the digital use cases. Despite the low performance for some cases in print-scan and print-scan compression images as noticed, one can note overall detection accuracy gain in cross-dataset testing while remaining comparable with the other SOTA algorithms. 

To conduct a comprehensive evaluation of the detection performance of the proposed method, we have benchmarked several existing SOTA targeting generalized S-MAD tasks. Besides constructing a representative morph database we are trying to simulate the operational application scenario, further we will submit the algorithms to third-party tests such as NIST (National Institute of Standards and Technology) Face Analysis Technology Evaluation (FATE) \cite{ngan2021face} or Bologna Online Evaluation Platform (BOEP) \cite{raja2020morphing}. In this work, the proposed method has not been submitted, but the selected reference algorithm based on hybrid features \cite{RagISBA2019} has been tested in FRVT MORPH and the performance is reported in \cite{ngan2021face}, and the Motimodality-based algorithm \cite{raghavendra2022multimodality} has been evaluated in BOEP.

Meanwhile, it should also be noted that in this work we only applied the vanilla Vision Transformer model pretrained with digital images on the image classification task. Hence it remains future works to plug in improved Vision Transformer models or replace the pretraining strategy with MAD-related tasks on different types of images. 

{
    \small
    \bibliographystyle{ieeenat_fullname}
    \bibliography{main}
}

\newpage
\appendix
\section{Appendix}
\subsection{Complete tables of MAD performances}
In this appendix, detailed results of quantitative MAD performance are included. Statistical analysis and discussions are in the main body of the paper. In each table, the intra-dataset testing case with the same training and testing dataset is marked with blue colour. Within each cross-dataset testing case (e.g., trained with Landmarks-I in the digital type of images, tested with Landmarks-II in the digital type of images), the best-performed value is marked in bold. 

\begin{table*}[hbp]
  \centering
  \caption{Quantitative performance of MAD on FRGC morph database. Training morphing type: Landmarks-I \cite{raghavendra2017face}.}
  \resizebox{1\linewidth}{!}{
    \begin{tabular}{|c|c|c|c|c|c|c|c|c|c|c|}
    \hline
     \textbf{Testing} & \multirow{3}[6]{*}{\textbf{MAD Algorithms}} & \multicolumn{3}{c|}{\textbf{Digital}} & \multicolumn{3}{c|}{\textbf{Print-scan}} & \multicolumn{3}{c|}{\textbf{Print-scan with compression}} \bigstrut\\
\cline{3-11}       \textbf{Morphing Type}         & \multicolumn{1}{c|}{} & \multicolumn{1}{c|}{\multirow{2}[4]{*}{D-EER(\%)}} & \multicolumn{2}{p{8.6em}|}{BPCER @ MACER =  } & \multicolumn{1}{c|}{\multirow{2}[4]{*}{D-EER(\%)}} & \multicolumn{2}{p{8.6em}|}{BPCER @ MACER =  } & \multicolumn{1}{c|}{\multirow{2}[4]{*}{D-EER(\%)}} & \multicolumn{2}{p{8.6em}|}{BPCER @ MACER =  } \bigstrut\\
\cline{4-5}\cline{7-8}\cline{10-11}       \multirow{3}[6]{*}{}         & \multicolumn{1}{c|}{} &       & 5.00\%   & 10.00\%  &       & 5.00\%   & 10.00\%  &       & 5.00\%   & 10.00\% \bigstrut\\
    \hline
     \cellcolor{mygray}
     & \cellcolor{mygray}Ensemble Features \cite{EnsembleFeatures_2020} & \cellcolor{mygray} \textbf{0.00}     & \cellcolor{mygray} \textbf{0.00}     & \cellcolor{mygray} \textbf{0.00}     & \cellcolor{mygray}2.35  & \cellcolor{mygray}1.45  & \cellcolor{mygray}0.96  & \cellcolor{mygray}2.58  & \cellcolor{mygray}1.71  & \cellcolor{mygray}1.54 \bigstrut\\
     
\cline{2-11}         \cellcolor{mygray}   &  \cellcolor{mygray}Hybrid Features \cite{RagISBA2019} & \cellcolor{mygray} 0.16  & \cellcolor{mygray} \textbf{0.00}     & \cellcolor{mygray} \textbf{0.00}     & \cellcolor{mygray} 1.85  & \cellcolor{mygray} 0.85  & \cellcolor{mygray} 0.34  & \cellcolor{mygray}2.25  & \cellcolor{mygray}1.12  & \cellcolor{mygray} 0.51 \bigstrut\\

\cline{2-11}         \cellcolor{mygray}   &  \cellcolor{mygray}Deep Features \cite{raja2017transferable} & \cellcolor{mygray} \textbf{0.00}  & \cellcolor{mygray} \textbf{0.00}     & \cellcolor{mygray} \textbf{0.00}     & \cellcolor{mygray} 2.85  & \cellcolor{mygray} 1.25  & \cellcolor{mygray} 0.17  & \cellcolor{mygray}2.05  & \cellcolor{mygray}1.02  & \cellcolor{mygray} 0.17 \bigstrut\\

\cline{2-11}         \cellcolor{mygray}   &  \cellcolor{mygray}Steerable Features \cite{ramachandra2019detecting} & \cellcolor{mygray} 5.48  & \cellcolor{mygray} 6.12     & \cellcolor{mygray} 3.60   & \cellcolor{mygray} 23.59  & \cellcolor{mygray} 62.90  & \cellcolor{mygray} 46.48 & \cellcolor{mygray}27.14  & \cellcolor{mygray}71.52  & \cellcolor{mygray} 57.28 \bigstrut\\

\cline{2-11}      \multicolumn{1}{|c|}{\multirow{-1}{*}{\cellcolor{mygray}Landmarks-I \cite{raghavendra2017face}}}    & \cellcolor{mygray}Multi-Modality \cite{raghavendra2022multimodality}  & \cellcolor{mygray}\textbf{0.00}  & \cellcolor{mygray} \textbf{0.00}     & \cellcolor{mygray} \textbf{0.00}     & \cellcolor{mygray}\textbf{0.00}  & \cellcolor{mygray}\textbf{0.00}  & \cellcolor{mygray} 0.52  & \cellcolor{mygray} 1.42  & \cellcolor{mygray} 0.71  & \cellcolor{mygray} 0.17 \bigstrut\\

\cline{2-11}      \cellcolor{mygray}    & \cellcolor{mygray}Residual AutoEncoder \cite{raja2022towards}  & \cellcolor{mygray}\textbf{0.00}  & \cellcolor{mygray} \textbf{0.00}     & \cellcolor{mygray} \textbf{0.00}     & \cellcolor{mygray}0.86  & \cellcolor{mygray}\textbf{0.00}  & \cellcolor{mygray}\textbf{0.00}  & \cellcolor{mygray} 0.69  & \cellcolor{mygray} \textbf{0.00}  & \cellcolor{mygray} \textbf{0.00} \bigstrut\\

\cline{2-11}      \cellcolor{mygray}    & \cellcolor{mygray}Multi-level Deep Features \cite{venkatesh2022multi}  & \cellcolor{mygray}\textbf{0.00}  & \cellcolor{mygray} \textbf{0.00}     & \cellcolor{mygray} \textbf{0.00}     & \cellcolor{mygray}1.37  & \cellcolor{mygray}0.51  & \cellcolor{mygray}0.17  & \cellcolor{mygray} 1.53  & \cellcolor{mygray} 0.51  & \cellcolor{mygray} 0.17 \bigstrut\\

\cline{2-11}      \cellcolor{mygray}    & \cellcolor{mygray}\textbf{Proposed Method}  & \cellcolor{mygray}0.51  & \cellcolor{mygray} \textbf{0.00}     & \cellcolor{mygray} \textbf{0.00}     & \cellcolor{mygray}2.23  & \cellcolor{mygray}0.86  & \cellcolor{mygray} 0.52  & \cellcolor{mygray} 1.89  & \cellcolor{mygray} 1.03  & \cellcolor{mygray} 0.69 \bigstrut\\

\cline{1-11}          & Ensemble Features \cite{EnsembleFeatures_2020} & 49.55 & 92.22 & 88.85 & 41.93 & 81.45 & 76.25 & 42.15 & 83.88 & 77.64 \bigstrut\\
\cline{2-11}                & Hybrid Features \cite{RagISBA2019} & 49.16 & 99.31 & 97.59 & 44.17 & 86.48 & 80.24 & 46.49 & 88.38 & 81.95 \bigstrut\\

\cline{2-11}       &  Deep Features \cite{raja2017transferable} &  42.36  & 88.46     & 77.28     & 41.90  & 85.04  & 72.30  & 41.58  & 91.17  & 71.11 \bigstrut\\

\cline{2-11}           &  Steerable Features \cite{ramachandra2019detecting} & 64.20  & 99.48     & 98.45   & 50.00  & 93.89  & 88.13 & 47.39  & 92.14  & 83.68 \bigstrut\\

\cline{2-11}       \multicolumn{1}{|c|}{\multirow{-1}{*}{Landmarks-II \cite{UBO_Morphing_Tool}}}    & Multi-Modality \cite{raghavendra2022multimodality}  & 41.51  & 80.61     &  71.86     & 35.37  & 79.64 &  73.92  &  37.86  & 84.46  &  76.96 \bigstrut\\

\cline{2-11}         & Residual AutoEncoder \cite{raja2022towards}  & 44.86  &  82.82    &  82.82     & \textbf{29.65}  & 99.48  & 94.85  &30.63  &  97.94 &  87.97 \bigstrut\\

\cline{2-11}         & Multi-level Deep Features \cite{venkatesh2022multi}  & 26.08  &  59.86    &  46.31     & 30.22  & \textbf{65.44}  & \textbf{55.32}  & \textbf{30.07}  & \textbf{64.93} &  \textbf{54.51} \bigstrut\\

\cline{2-11}         & \textbf{Proposed Method}  & \textbf{23.50}  & \textbf{49.57}     & \textbf{40.99}     & 40.10  & 85.59  & 75.35  & 37.26  & 88.21  & 79.03\bigstrut\\

\cline{1-11}           & Ensemble Features \cite{EnsembleFeatures_2020} & 0.22  & \textbf{0.00}     & \textbf{0.00}     & 13.36 & 27.44 & 16.46 & 14.77 & 27.27 & 19.38 \bigstrut\\
\cline{2-11}                 & Hybrid Features \cite{RagISBA2019} & 0.16  & \textbf{0.00}     & \textbf{0.00}     & 44.96 & 83.70  & 75.47 & 9.44  & 14.57 & 9.14 \bigstrut\\

\cline{2-11}       &  Deep Features \cite{raja2017transferable} &  0.16  & \textbf{0.00}     & \textbf{0.00}     & 1.08  & 0.17  &  \textbf{0.00}  &8.92  & 12.00  & 8.17 \bigstrut\\

\cline{2-11}           &  Steerable Features \cite{ramachandra2019detecting} & 7.16  & 9.77     & 4.63   & 3.46  & 2.42  & 1.54 & 38.91  & 87.30  & 79.58 \bigstrut\\

\cline{2-11}       \multicolumn{1}{|c|}{\multirow{-1}{*}{StyleGAN \cite{MorphStyleGAN2020}}}    & Multi-Modality \cite{raghavendra2022multimodality}  & \textbf{0.00}  & \textbf{0.00}      &  \textbf{0.00}      & \textbf{0.00}   & \textbf{0.00}   &  \textbf{0.00}   & 5.34  & \textbf{5.35}  &  3.21 \bigstrut\\

\cline{2-11}         & Residual AutoEncoder \cite{raja2022towards}  & 0.17  &  \textbf{0.00}    &  \textbf{0.00}     & 6.36  & 8.76  & 4.64  &6.03  &  6.19 &  3.78 \bigstrut\\

\cline{2-11}         & Multi-level Deep Features \cite{venkatesh2022multi}  & 0.32  &  \textbf{0.00}    &  \textbf{0.00}     & 3.95  & 2.05  & 0.51  & \textbf{5.32}  & 5.83 &  \textbf{2.74} \bigstrut\\

\cline{2-11}            & \textbf{Proposed Method}  & 2.57  & 1.37     & 0.34     & 12.35  & 23.16  & 14.75  & 12.35  & 25.56  & 14.07\bigstrut\\

\cline{1-11}           & Ensemble Features \cite{EnsembleFeatures_2020}& 39.16 & 73.14 & 65.35 & 9.45  & 14.57 & 8.74  & 8.95  & 15.26 & 9.26 \bigstrut\\
\cline{2-11}                 & Hybrid Features \cite{RagISBA2019} & 46.82 & 86.62 & 81.64 & 12.32 & 19.72 & 13.20  & 9.74  & 15.95 & 8.91 \bigstrut\\

\cline{2-11}       &  Deep Features \cite{raja2017transferable} &  44.94  & 84.59    & 74.47     & 3.73  & 2.40  &  1.02  & 23.83  & 66.55  & 49.22 \bigstrut\\

\cline{2-11}           &  Steerable Features \cite{ramachandra2019detecting} & 50.00  & 99.82     & 97.59   & 3.26  & 2.22  & 2.15 & 28.82  & 79.24  & 60.20 \bigstrut\\

\cline{2-11}       \multicolumn{1}{|c|}{\multirow{-1}{*}{MIPGAN-I \cite{zhang2021mipgan}}}    & Multi-Modality \cite{raghavendra2022multimodality}  & 41.39  & 78.79      &  72.55      & \textbf{0.00}   & \textbf{0.00}  & \textbf{0.00}   & \textbf{8.14}  & \textbf{10.89}  &  \textbf{7.14} \bigstrut\\
\cline{2-11}         & Residual AutoEncoder \cite{raja2022towards}  & 16.42  &  29.21    &  20.45     & 33.33  & 96.05  & 90.03  &25.49  &  98.11 &  89.00\bigstrut\\

\cline{2-11}         & Multi-level Deep Features \cite{venkatesh2022multi}  & 20.97  &  46.65    &  34.13     & 11.35  & 29.84  & 14.57  &23.47  & 62.60 &  47.34 \bigstrut\\

\cline{2-11}             & \textbf{Proposed Method}  & \textbf{11.15}  & \textbf{23.16}     & \textbf{12.35}     & 19.38  & 40.31  & 28.82  & 27.62  & 62.95  & 48.89\bigstrut\\

\cline{1-11}         & Ensemble Features \cite{EnsembleFeatures_2020} & 34.13 & 70.49 & 61.57 & 5.32  & 6.68  & 2.57  & 6.72  & 8.16  & 4.14 \bigstrut\\
\cline{2-11}                 & Hybrid Features \cite{RagISBA2019} & 44.96 & 83.7  & 75.47 & 5.90   & 8.42  & 3.23  & 5.67  & 6.18  & 2.91 \bigstrut \\

\cline{2-11}       &  Deep Features \cite{raja2017transferable} &  40.37  & 60.32     & 50.36     & 1.36  & \textbf{0.00}  &  \textbf{0.00}  & 8.37  & 24.69  & 16.17 \bigstrut\\

\cline{2-11}           &  Steerable Features \cite{ramachandra2019detecting} & 50.00  & 98.77     & 97.25   & 6.84  & 7.54  & 6.12 & 31.16  & 85.42  & 73.75 \bigstrut\\

\cline{2-11}       \multicolumn{1}{|c|}{\multirow{-1}{*}{MIPGAN-II \cite{zhang2021mipgan}}}    & Multi-Modality \cite{raghavendra2022multimodality}  & 35.86  & 66.72      &  56.43      & \textbf{0.00}   & \textbf{0.00}  & \textbf{0.00}   & \textbf{4.47}  & \textbf{3.57}  & 0.35 \bigstrut\\
\cline{2-11}         & Residual AutoEncoder \cite{raja2022towards}  & 13.92  &  21.82    &  17.01     & 8.80  & 77.66  & 1.03  &5.60  &  8.06 & \textbf{0.17}\bigstrut\\

\cline{2-11}         & Multi-level Deep Features \cite{venkatesh2022multi}  & 19.20  &  39.79    &  29.50     & 3.95  & 2.05  & 0.51  &8.73  & 26.07 &  6.34 \bigstrut\\

    \cline{2-11}             & \textbf{Proposed Method}  & \textbf{8.40}  & \textbf{12.52}    & \textbf{6.35}     & 5.15  & 5.49  & 0.86  & 8.58  & 24.01  & 3.77\bigstrut\\

    \hline
    \end{tabular}%
    }
  \label{tab:MADlandmark1}%
\end{table*}%


\begin{table*}[h!]
  \centering
  \caption{Quantitative performance of MAD on FRGC morph database. Training morphing type: Landmarks-II \cite{UBO_Morphing_Tool}.}
  \resizebox{1\linewidth}{!}{
    \begin{tabular}{|c|c|c|c|c|c|c|c|c|c|c|}
    \hline
     \textbf{Testing} & \multirow{3}[6]{*}{\textbf{MAD Algorithms}} & \multicolumn{3}{c|}{\textbf{Digital}} & \multicolumn{3}{c|}{\textbf{Print-scan}} & \multicolumn{3}{c|}{\textbf{Print-scan with compression}} \bigstrut\\
\cline{3-11}      \textbf{Morphing Type}         & \multicolumn{1}{c|}{} & \multicolumn{1}{c|}{\multirow{2}[4]{*}{D-EER(\%)}} & \multicolumn{2}{c|}{BPCER @ MACER =  } & \multicolumn{1}{c|}{\multirow{2}[4]{*}{D-EER(\%)}} & \multicolumn{2}{c|}{BPCER @ MACER =  } & \multicolumn{1}{c|}{\multirow{2}[4]{*}{D-EER(\%)}} & \multicolumn{2}{c|}{BPCER @ MACER =  } \bigstrut\\
\cline{4-5}\cline{7-8}\cline{10-11}      \multirow{3}[6]{*}{}          & \multicolumn{1}{c|}{} &       & 5.00\%   & 10.00\%  &       & 5.00\%   & 10.00\%  &       & 5.00\%   & 10.00\% \bigstrut\\
    \hline
      & Ensemble Features \cite{EnsembleFeatures_2020} & 48.57 & 97.77 & 95.36 & 24.19 & 52.48 & 43.22 & 21.64 & 47.51 & 36.19 \bigstrut\\
\cline{2-11}              & Hybrid Features \cite{RagISBA2019} & 45.67 & 96.91 & 94.16 & 32.26 & 77.87 & 66.55 & 24.51 & 50.94 & 40.65 \bigstrut\\

\cline{2-11}       &  Deep Features \cite{raja2017transferable} &  47.22  & 89.94     & 68.98     & 26.66  & 55.40  &  42.53  & 25.24  & 51.80  & 40.99 \bigstrut\\

\cline{2-11}           &  Steerable Features \cite{ramachandra2019detecting} & 50.00  & 95.12     & 91.25  & 37.72 & 93.48  & 82.84 & 37.13  & 92.62  & 83.71 \bigstrut\\

\cline{2-11}       \multicolumn{1}{|c|}{\multirow{-1}{*}{Landmarks-I \cite{raghavendra2017face}}}    & Multi-Modality \cite{raghavendra2022multimodality}  & 19.54  & 42.36      &  31.21      & 34.29   & 84.28  & 73.32   & 38.19  & 81.42  & 71.25\bigstrut\\
\cline{2-11}         & Residual AutoEncoder \cite{raja2022towards}  & \textbf{3.59}  &  \textbf{2.58}    &  \textbf{1.72}     & \textbf{9.85}  & \textbf{19.97}  & \textbf{9.72}  &\textbf{13.82}  &  \textbf{44.44} & \textbf{21.01}\bigstrut\\

\cline{2-11}         & Multi-level Deep Features \cite{venkatesh2022multi}  & 14.92  &  33.10    &  22.64     & 29.40  & 61.57  & 51.11  &24.02  & 54.71 &  44.94 \bigstrut\\

\cline{2-11}           & \textbf{Proposed Method}  & 14.92  & 33.10     & 22.64     & 37.29  & 80.41  & 67.01  & 26.63  & 63.75  & 52.92\bigstrut\\

\cline{1-11}           \cellcolor{mygray} & \cellcolor{mygray}Ensemble Features \cite{EnsembleFeatures_2020} & \cellcolor{mygray}3.62  & \cellcolor{mygray}2.22  & \cellcolor{mygray}0.68  & \cellcolor{mygray}6.32  & \cellcolor{mygray}7.97  & \cellcolor{mygray}2.42  & \cellcolor{mygray}5.57  & \cellcolor{mygray}6.41  & \cellcolor{mygray}2.42 \bigstrut\\
\cline{2-11}     \cellcolor{mygray}    & \cellcolor{mygray}Hybrid Features \cite{RagISBA2019} & \cellcolor{mygray} 1.53  & \cellcolor{mygray} 0.17  & \cellcolor{mygray} \textbf{0.00}     & \cellcolor{mygray} 5.21  & \cellcolor{mygray} 5.19  & \cellcolor{mygray} 3.14  & \cellcolor{mygray} 5.37  & \cellcolor{mygray} 5.71  & \cellcolor{mygray} 3.46 \bigstrut\\

\cline{2-11}    \cellcolor{mygray}     & \cellcolor{mygray}Deep Features \cite{raja2017transferable}  & \cellcolor{mygray} 6.16  &  \cellcolor{mygray}6.51    &  \cellcolor{mygray}3.94     & \cellcolor{mygray}6.65 & \cellcolor{mygray}9.94  & \cellcolor{mygray}4.88  &\cellcolor{mygray}7.13  &  \cellcolor{mygray}12.50 & \cellcolor{mygray}5.55\bigstrut\\

\cline{2-11}    \cellcolor{mygray}     & \cellcolor{mygray} Steerable Features \cite{ramachandra2019detecting}  & \cellcolor{mygray}27.78  &  \cellcolor{mygray}69.46    &  \cellcolor{mygray}53.68     & \cellcolor{mygray}30.55  & \cellcolor{mygray}79.75  & \cellcolor{mygray}68.05  &\cellcolor{mygray}29.54  &  \cellcolor{mygray}76.41 & \cellcolor{mygray}63.19\bigstrut\\

\cline{2-11}       \multicolumn{1}{|c|}{\multirow{-1}{*}{\cellcolor{mygray}Landmarks-II \cite{UBO_Morphing_Tool}}}    & \cellcolor{mygray}Multi-Modality \cite{raghavendra2022multimodality}  & \cellcolor{mygray} \textbf{0.00}  & \cellcolor{mygray} \textbf{0.00}      &  \cellcolor{mygray} \textbf{0.00}      & \cellcolor{mygray} \textbf{1.61}   & \cellcolor{mygray} \textbf{0.53}  & \cellcolor{mygray} \textbf{0.35}   & \cellcolor{mygray} \textbf{1.82}  & \cellcolor{mygray} \textbf{0.35}  & \cellcolor{mygray} \textbf{0.17}\bigstrut\\
\cline{2-11}    \cellcolor{mygray}     & \cellcolor{mygray}Residual AutoEncoder \cite{raja2022towards}  & \cellcolor{mygray}2.58  &  \cellcolor{mygray}2.06    &  \cellcolor{mygray}1.72     & \cellcolor{mygray}7.99  & \cellcolor{mygray}12.15  & \cellcolor{mygray}5.73  &\cellcolor{mygray}9.38  &  \cellcolor{mygray}19.97 & \cellcolor{mygray}9.2\bigstrut\\

\cline{2-11}   \cellcolor{mygray}     & \cellcolor{mygray}Multi-level Deep Features \cite{venkatesh2022multi}  & \cellcolor{mygray}10.63  & \cellcolor{mygray}25.21     & \cellcolor{mygray}11.66     & \cellcolor{mygray}5.39  & \cellcolor{mygray}5.93  & \cellcolor{mygray} 2.09 & \cellcolor{mygray}6.19  & \cellcolor{mygray}7.46  & \cellcolor{mygray}3.29\bigstrut\\

\cline{2-11}   \cellcolor{mygray}     & \cellcolor{mygray}\textbf{Proposed Method}  & \cellcolor{mygray}10.63  & \cellcolor{mygray}25.21     & \cellcolor{mygray}11.66     & \cellcolor{mygray}11.46  & \cellcolor{mygray}19.62  & \cellcolor{mygray} 12.15 & \cellcolor{mygray}10.92  & \cellcolor{mygray}25.65  & \cellcolor{mygray}11.96\bigstrut\\

\cline{1-11}    & Ensemble Features \cite{EnsembleFeatures_2020} & 29.67 & 61.92 & 52.48 & 27.18 & 61.57 & 50.60  & 29.18 & 62.14 & 52.48 \bigstrut\\
\cline{2-11}                & Hybrid Features \cite{RagISBA2019} & 34.76 & 74.44 & 62.95 & 34.80  & 67.23 & 58.14 & 23.17 & 49.22 & 38.25 \bigstrut\\

\cline{2-11}       &  Deep Features \cite{raja2017transferable} &  30.13  & 35.19     & 40.29     & 1.37  & 0.17  &  0.17  & 26.65  & 57.10  & 46.31 \bigstrut\\

\cline{2-11}           &  Steerable Features \cite{ramachandra2019detecting} & 50.00  & 98.28     & 96.56   & 1.21  & 0.34  & \textbf{0.00} & 35.52  & 85.31  & 50 \bigstrut\\

\cline{2-11}       \multicolumn{1}{|c|}{\multirow{-1}{*}{StyleGAN \cite{MorphStyleGAN2020}}}    & Multi-Modality \cite{raghavendra2022multimodality}  & 25.4  & 62.09      &  49.91      & \textbf{0.16}   & \textbf{0.00}  & \textbf{0.00}   & 15.53  & 31.96  & 23.21\bigstrut\\
\cline{2-11}         & Residual AutoEncoder \cite{raja2022towards}  & \textbf{9.15}  &  \textbf{16.84}    & \textbf{8.59}    & 8.16  & 13.19  & 6.25  & \textbf{8.09}  &  \textbf{11.81} & \textbf{6.25}\bigstrut\\

\cline{2-11}         & Multi-level Deep Features \cite{venkatesh2022multi}  & 22.64  &  51.97    &  35.51     & 4.64  & 4.45  & 1.37  &25.55  & 51.28 &  42.53 \bigstrut\\

\cline{2-11}           & \textbf{Proposed Method}  & 22.64  & 51.97     & 35.51     & 25.21  & 60.72  & 45.97  & 34.48  & 73.07  & 60.38\bigstrut\\

\cline{1-11}         & Ensemble Features \cite{EnsembleFeatures_2020} & 30.23 & 65.35 & 53.17 & 43.92 & 87.65 & 79.24 & 44.24 & 89.23 & 82.33 \bigstrut\\
\cline{2-11}               & Hybrid Features \cite{RagISBA2019} & 46.29 & 84.04 & 77.01 & 34.16 & 71.18 & 64.66 & 35.50  & 76.84 & 65.52 \bigstrut\\

\cline{2-11}       &  Deep Features \cite{raja2017transferable} &  38.18  & 77.53     & 67.92     & 1.20  & 0.34  &  0.17  & 27.41  & 65.69  & 52.65 \bigstrut\\

\cline{2-11}           &  Steerable Features \cite{ramachandra2019detecting} & 36.51  & 86.96     & 76.51   & 2.61 & 0.85  & \textbf{0.00} & 32.73  & 87.13  & 74.19 \bigstrut\\

\cline{2-11}       \multicolumn{1}{|c|}{\multirow{-1}{*}{MIPGAN-I \cite{zhang2021mipgan}}}    & Multi-Modality \cite{raghavendra2022multimodality}  & 24.75  & 72.14     &  42.88      & \textbf{0.37}   & \textbf{0.00}  & \textbf{0.00}   & 40.51  & 82.14  & 73.92\bigstrut\\
\cline{2-11}         & Residual AutoEncoder \cite{raja2022towards}  & \textbf{12.89}  & \textbf{22.51}    &  \textbf{15.46}     & 6.45  & 7.81  & 3.82  &\textbf{1.44}  &  \textbf{0.52} & \textbf{0.17}\bigstrut\\

\cline{2-11}         & Multi-level Deep Features \cite{venkatesh2022multi}  & 23.67  &  55.75    &  42.71     & 4.10  & 3.08 & 0.51  &28.58  & 62.26 &  51.80\bigstrut\\

\cline{2-11}            & \textbf{Proposed Method}  & 23.67  & 55.75     & 42.71     & 19.04  & 47.86  & 36.36  & 23.16  & 57.29  &45.80\bigstrut\\

\cline{1-11}          & Ensemble Features \cite{EnsembleFeatures_2020} & 27.13 & 58.83 & 45.45 & 33.57 & 77.35 & 65.52 & 40.46 & 84.9  & 75.47 \bigstrut\\
\cline{2-11}                & Hybrid Features \cite{RagISBA2019} & 46.82 & 83.53 & 75.81 & 35.91 & 77.18 & 65.24 & 36.50  & 79.24 & 68.78 \bigstrut\\

\cline{2-11}       &  Deep Features \cite{raja2017transferable} & 36.71  & 78.73     & 68.61     & 1.04  & \textbf{0.00}  &  \textbf{0.00}  & 34.40  & 71.18  & 59.69 \bigstrut\\

\cline{2-11}           &  Steerable Features \cite{ramachandra2019detecting} & 36.87  & 90.33     & 79.41   & 3.19  & 1.88& 0.34 & 34.12  & 89.19  & 78.91 \bigstrut\\

\cline{2-11}       \multicolumn{1}{|c|}{\multirow{-1}{*}{MIPGAN-II \cite{zhang2021mipgan}}}    & Multi-Modality \cite{raghavendra2022multimodality}  & 20.92  & 33.27      &  43.05      & \textbf{0.91}   & \textbf{0.00}  & \textbf{0.00}   & 33.33  &77.85  & 67.5\bigstrut\\

\cline{2-11}         & Residual AutoEncoder \cite{raja2022towards}  & \textbf{8.93}  &  \textbf{14.43}    &  \textbf{8.59}     & 4.51  &4.51  & 2.60  &\textbf{2.26}  &  \textbf{1.04} & \textbf{0.17}\bigstrut\\

\cline{2-11}         & Multi-level Deep Features \cite{venkatesh2022multi}  & 29.67  &  67.92    &  51.11     & 1.73  & \textbf{0.00}  & \textbf{0.00} &34.46  & 68.95 &  58.14 \bigstrut\\

\cline{2-11}           & \textbf{Proposed Method}  & 29.67  & 67.92     & 51.11     & 30.53  & 78.22  & 63.46  & 28.47  & 68.95  & 55.06\bigstrut\\

    \hline
    \end{tabular}%
    }
  \label{tab:MADlandmark2}%
\end{table*}%

\begin{table*}[h!]
  \centering
  \caption{Quantitative performance of MAD on FRGC morph database. Training morphing type: StyleGAN \cite{MorphStyleGAN2020}.}
  \resizebox{1\linewidth}{!}{
    \begin{tabular}{|c|c|c|c|c|c|c|c|c|c|c|}
    \hline
     \textbf{Testing} & \multirow{3}[6]{*}{\textbf{MAD Algorithms}} & \multicolumn{3}{c|}{\textbf{Digital}} & \multicolumn{3}{c|}{\textbf{Print-scan}} & \multicolumn{3}{c|}{\textbf{Print-scan with compression}} \bigstrut\\
\cline{3-11}         \textbf{Morphing Type}         & \multicolumn{1}{|c|}{} & \multicolumn{1}{c|}{\multirow{2}[4]{*}{D-EER(\%)}} & \multicolumn{2}{c|}{BPCER @ MACER =  } & \multicolumn{1}{c|}{\multirow{2}[4]{*}{D-EER(\%)}} & \multicolumn{2}{c|}{BPCER @ MACER =  } & \multicolumn{1}{c|}{\multirow{2}[4]{*}{D-EER(\%)}} & \multicolumn{2}{c|}{BPCER @ MACER =  } \bigstrut\\
\cline{4-5}\cline{7-8}\cline{10-11}         \multirow{3}[6]{*}{}       & \multicolumn{1}{c|}{} &       & 5.00\%   & 10.00\%  &       & 5.00\%   & 10.00\%  &       & 5.00\%   & 10.00\% \bigstrut\\
    \hline
      & Ensemble Features \cite{EnsembleFeatures_2020} & 0.32  & \textbf{0.00}     & \textbf{0.00}     & 16.60  & 28.13 & 19.89 & 13.89 & 22.12 & 17.66 \bigstrut\\
\cline{2-11}                & Hybrid Features \cite{RagISBA2019} & 0.42  & \textbf{0.00}     & \textbf{0.00}     & 15.26 & \textbf{26.41} & \textbf{17.66} & 14.37 & 22.81 & 16.92 \bigstrut\\

\cline{2-11}       &  Deep Features \cite{raja2017transferable} &  0.16  & \textbf{0.00}     & \textbf{0.00}     & 24.67  & 55.74  &  41.80  & 13.36  & 34.30  & 18.69 \bigstrut\\

\cline{2-11}           &  Steerable Features \cite{ramachandra2019detecting} & 6.17  & 7.71     & 3.94   & 33.92  & 81.81  & 69.46 & 35.62  & 83.87  & 74.19 \bigstrut\\

\cline{2-11}       \multicolumn{1}{|c|}{\multirow{-1}{*}{Landmarks-I \cite{raghavendra2017face}}}    & Multi-Modality \cite{raghavendra2022multimodality}  & \textbf{0.00}  & \textbf{0.00}      &  \textbf{0.00}      & 20.00   & 41.96  & 29.64   & 11.97  & 17.85  & 12.85\bigstrut\\

\cline{2-11}         & Residual AutoEncoder \cite{raja2022towards}  & 0.17  &  \textbf{0.00}    &  \textbf{0.00}     & 15.02  & 36.08  & 22.34  &8.93  &  17.35 & 7.90\bigstrut\\

\cline{2-11}         & Multi-level Deep Features \cite{venkatesh2022multi}  & 0.16  &  \textbf{0.00}    &  \textbf{0.00}    & \textbf{1.37}  & \textbf{0.51}  & \textbf{0.17} & \textbf{1.53}  & \textbf{0.51} &  \textbf{0.17} \bigstrut\\

\cline{2-11}            & \textbf{Proposed Method}  & 5.15  & 5.15     & 2.74     & 14.95  & 30.24  & 21.31  & 12.71  & 24.74  & 15.81\bigstrut\\

\cline{1-11}           & Ensemble Features \cite{EnsembleFeatures_2020} & 44.72 & 89.53 & 80.61 & 38.31 & 78.50  & 69.15 & 38.84 & 83.70  & 74.17 \bigstrut\\
\cline{2-11}               & Hybrid Features \cite{RagISBA2019} & 45.65 & 90.22 & 84.56 & 34.18 & 81.95 & 70.53 & 32.93 & 78.50  & 64.12 \bigstrut\\

\cline{2-11}       &  Deep Features \cite{raja2017transferable} &  43.43  & 95.31     & 83.12     & 45.34  & 96.10  &  88.48  & 30.19  & 65.45  & 54.68 \bigstrut\\

\cline{2-11}           &  Steerable Features \cite{ramachandra2019detecting} & 50.00  & 99.65     & 99.10   & 40.81  & 83.42  & 72.60 & 39.13  & 84.89  & 74.82 \bigstrut\\

\cline{2-11}       \multicolumn{1}{|c|}{\multirow{-1}{*}{Landmarks-II \cite{UBO_Morphing_Tool}}}    & Multi-Modality \cite{raghavendra2022multimodality}  & 42.69  & 88.50      &  78.38      & \textbf{26.72}   & \textbf{73.21}  & \textbf{56.25}   & 27.87  & 68.75  & 57.14\bigstrut\\

\cline{2-11}         & Residual AutoEncoder \cite{raja2022towards}  & 38.94  &  74.40    &  \textbf{58.25}     & 34.72  & 99.66  & 93.13  & \textbf{27.32}  &  92.10 & 83.85\bigstrut\\

\cline{2-11}         & Multi-level Deep Features \cite{venkatesh2022multi}  & \textbf{32.17}  &  \textbf{69.46}    &  60.89     & 30.22  & 65.44  & 55.32 &30.07  & \textbf{64.93} &  \textbf{54.51} \bigstrut\\

\cline{2-11}            & \textbf{Proposed Method}  & 37.91  & 87.48     & 77.02     & 43.23  & 92.53  & 85.07  & 40.73  & 88.03  & 78.51\bigstrut\\

\cline{1-11}           \cellcolor{mygray} & \cellcolor{mygray}Ensemble Features \cite{EnsembleFeatures_2020} &\cellcolor{mygray} \textbf{0.00}     &\cellcolor{mygray} \textbf{0.00}     &\cellcolor{mygray} \textbf{0.00}     & \cellcolor{mygray} \textbf{0.00}     &\cellcolor{mygray} \textbf{0.00}     & \cellcolor{mygray} \textbf{0.00}     & \cellcolor{mygray} \textbf{0.00}     & \cellcolor{mygray} \textbf{0.00}     & \cellcolor{mygray} \textbf{0.00} \bigstrut\\
\cline{2-11}           \cellcolor{mygray}      & \cellcolor{mygray}Hybrid Features \cite{RagISBA2019} & \cellcolor{mygray} \textbf{0.00}     & \cellcolor{mygray} \textbf{0.00}     & \cellcolor{mygray} \textbf{0.00}     & \cellcolor{mygray} \textbf{0.00}     &\cellcolor{mygray} \textbf{0.00}     & \cellcolor{mygray} \textbf{0.00}     &\cellcolor{mygray} \textbf{0.00}     &\cellcolor{mygray} \textbf{0.00}     & \cellcolor{mygray} \textbf{0.00} \bigstrut\\

\cline{2-11}         \cellcolor{mygray}   &  \cellcolor{mygray}Deep Features \cite{raja2017transferable} & \cellcolor{mygray} \textbf{0.00}  & \cellcolor{mygray} \textbf{0.00}     & \cellcolor{mygray} \textbf{0.00}     & \cellcolor{mygray} \textbf{0.00}  & \cellcolor{mygray} \textbf{0.00}  & \cellcolor{mygray} \textbf{0.00}  & \cellcolor{mygray}\textbf{0.00}  & \cellcolor{mygray}\textbf{0.00}  & \cellcolor{mygray} \textbf{0.00} \bigstrut\\

\cline{2-11}         \cellcolor{mygray}   &  \cellcolor{mygray}Steerable Features \cite{ramachandra2019detecting} & \cellcolor{mygray}3.75  & \cellcolor{mygray} 3.43     & \cellcolor{mygray} 1.54   & \cellcolor{mygray} \textbf{0.00}  & \cellcolor{mygray} \textbf{0.00}  & \cellcolor{mygray}\textbf{0.00} & \cellcolor{mygray}20.66  & \cellcolor{mygray}56.69  & \cellcolor{mygray} 43.39 \bigstrut\\

\cline{2-11}      \multicolumn{1}{|c|}{\multirow{-1}{*}{\cellcolor{mygray}StyleGAN \cite{MorphStyleGAN2020}}}    & \cellcolor{mygray}Multi-Modality \cite{raghavendra2022multimodality}  & \cellcolor{mygray}\textbf{0.00}  & \cellcolor{mygray} \textbf{0.00}     & \cellcolor{mygray} \textbf{0.00}     & \cellcolor{mygray}\textbf{0.00}  & \cellcolor{mygray}\textbf{0.00}  & \cellcolor{mygray} \textbf{0.00}  & \cellcolor{mygray} \textbf{0.00} & \cellcolor{mygray} \textbf{0.00}  & \cellcolor{mygray}\textbf{0.00} \bigstrut\\

\cline{2-11}      \cellcolor{mygray}    & \cellcolor{mygray}Residual AutoEncoder \cite{raja2022towards}  & \cellcolor{mygray}\textbf{0.00}  & \cellcolor{mygray} \textbf{0.00}     & \cellcolor{mygray} \textbf{0.00}     & \cellcolor{mygray}0.08  & \cellcolor{mygray}\textbf{0.00}  & \cellcolor{mygray}\textbf{0.00}  & \cellcolor{mygray} 0.34  & \cellcolor{mygray} \textbf{0.00}  & \cellcolor{mygray} \textbf{0.00} \bigstrut\\

\cline{2-11}      \cellcolor{mygray}    & \cellcolor{mygray}Multi-level Deep Features \cite{venkatesh2022multi}  & \cellcolor{mygray}\textbf{0.00}  & \cellcolor{mygray} \textbf{0.00}     & \cellcolor{mygray} \textbf{0.00}     & \cellcolor{mygray}3.95  & \cellcolor{mygray}2.05  & \cellcolor{mygray}0.51  & \cellcolor{mygray} 5.32 & \cellcolor{mygray} 5.83  & \cellcolor{mygray} 2.74 \bigstrut\\

\cline{2-11}         \cellcolor{mygray}    & \cellcolor{mygray}\textbf{Proposed Method}  & \cellcolor{mygray}\textbf{0.00}  & \cellcolor{mygray} \textbf{0.00}     & \cellcolor{mygray} \textbf{0.00}     & \cellcolor{mygray}1.03  & \cellcolor{mygray}0.17  & \cellcolor{mygray} \textbf{0.00} & \cellcolor{mygray}1.20  & \cellcolor{mygray} \textbf{0.00}  & \cellcolor{mygray} \textbf{0.00} \bigstrut\\

\cline{1-11}            & Ensemble Features \cite{EnsembleFeatures_2020} & 39.97 & 75.98 & 68.78 & 20.21 & 42.14 & 33.44 & 20.73 & 45.28 & 36.53 \bigstrut\\
\cline{2-11}                & Hybrid Features \cite{RagISBA2019} & 46.45 & 86.79 & 77.87 & 29.34 & 59.19 & 47.51 & 24.87 & 51.62 & 41.18 \bigstrut\\

\cline{2-11}       &  Deep Features \cite{raja2017transferable} &  23.95  & 48.19     & 38.25     & \textbf{0.00}  & \textbf{0.00}  &  \textbf{0.00}  & 26.40  & 63.97  & 44.57 \bigstrut\\

\cline{2-11}           &  Steerable Features \cite{ramachandra2019detecting} & 50.00  & 98.45     & 96.91   & \textbf{0.00}  & \textbf{0.00}  & \textbf{0.00} & 31.76  & 82.33  & 69.46 \bigstrut\\

\cline{2-11}       \multicolumn{1}{|c|}{\multirow{-1}{*}{MIPGAN-I \cite{zhang2021mipgan}}}    & Multi-Modality \cite{raghavendra2022multimodality}  & 33.73  & 69.53      &  61.69     & \textbf{0.00}   & \textbf{0.00}  & \textbf{0.00}   & \textbf{20.07}  & \textbf{39.46}  & \textbf{30.39}\bigstrut\\

\cline{2-11}         & Residual AutoEncoder \cite{raja2022towards}  & \textbf{16.61}  &  \textbf{27.66}    &  \textbf{20.45}     & 71.95  & 99.83  & 99.66  &41.20  &  81.44 & 75.95\bigstrut\\

\cline{2-11}         & Multi-level Deep Features \cite{venkatesh2022multi}  & 27.21  &  62.43    &  48.19     & 11.35 & 29.54  & 14.57 &23.47  & 62.60 &  47.34 \bigstrut\\

\cline{2-11}             & \textbf{Proposed Method}  & 21.61  & 49.91     & 35.68     & 37.56  & 79.07  & 72.90  & 32.76  & 65.87  & 56.43\bigstrut\\

\cline{1-11}            & Ensemble Features \cite{EnsembleFeatures_2020} & 39.93 & 73.58 & 66.89 & 15.78 & 28.14 & 19.38 & 13.72 & 28.98 & 16.63 \bigstrut\\
\cline{2-11}                 & Hybrid Features \cite{RagISBA2019} & 44.72 & 82.16 & 73.75 & 19.36 & 43.22 & 28.64 & 16.98 & 32.93 & 23.84 \bigstrut\\

\cline{2-11}       &  Deep Features \cite{raja2017transferable} &  42.19  & 70.42     & 60.48     & \textbf{0.00}  & \textbf{0.00}  &  \textbf{0.00}  & 9.98 & 20.41  & 9.94 \bigstrut\\

\cline{2-11}           &  Steerable Features \cite{ramachandra2019detecting} & 50.00  & 98.28     & 95.38   & \textbf{0.00}  & \textbf{0.00}  &\textbf{0.00} & 31.07 & 79.41  & 63.46 \bigstrut\\

\cline{2-11}       \multicolumn{1}{|c|}{\multirow{-1}{*}{MIPGAN-II \cite{zhang2021mipgan}}}    & Multi-Modality \cite{raghavendra2022multimodality}  & 37.56  & 75.98      & 65.86      & \textbf{0.00}   & \textbf{0.00}  & \textbf{0.00}   & \textbf{7.85}  & \textbf{12.67}  & \textbf{5.17}\bigstrut\\

\cline{2-11}         & Residual AutoEncoder \cite{raja2022towards}  & \textbf{15.09}  &  \textbf{26.63}    &  \textbf{19.76}    & 10.85  & 95.53  & 62.89  &11.53  &  58.42 & 21.82\bigstrut\\

\cline{2-11}         & Multi-level Deep Features \cite{venkatesh2022multi}  & 25.04  &  57.46    &  45.28    & 3.06 & 0.51  & 0.17 &8.73  & 26.07 &  6.34 \bigstrut\\

\cline{2-11}             & \textbf{Proposed Method}  & 18.01  & 44.43     & 32.42     & 13.72  & 64.67  & 30.02  & 12.52  & 38.94  & 17.67\bigstrut\\

    \hline
    \end{tabular}%
    }
  \label{tab:MADstylegan}%
\end{table*}%


\begin{table*}[h!]
  \centering
  \caption{Quantitative performance of MAD on FRGC morph database. Training morphing type: MIPGAN-I \cite{zhang2021mipgan}.}
  \resizebox{1\linewidth}{!}{
    \begin{tabular}{|c|c|c|c|c|c|c|c|c|c|c|}
    \hline
     \textbf{Testing} & \multirow{3}[6]{*}{\textbf{MAD Algorithms}} & \multicolumn{3}{c|}{\textbf{Digital}} & \multicolumn{3}{c|}{\textbf{Print-scan}} & \multicolumn{3}{c|}{\textbf{Print-scan with compression}} \bigstrut\\
\cline{3-11}           \textbf{Morphing Type}       & \multicolumn{1}{c|}{} & \multicolumn{1}{c|}{\multirow{2}[4]{*}{D-EER(\%)}} & \multicolumn{2}{c|}{BPCER @ MACER =  } & \multicolumn{1}{c|}{\multirow{2}[4]{*}{D-EER(\%)}} & \multicolumn{2}{c|}{BPCER @ MACER =  } & \multicolumn{1}{c|}{\multirow{2}[4]{*}{D-EER(\%)}} & \multicolumn{2}{c|}{BPCER @ MACER =  } \bigstrut\\
\cline{4-5}\cline{7-8}\cline{10-11}        \multirow{3}[6]{*}{}         & \multicolumn{1}{|c|}{} &       & 5.00\%   & 10.00\%  &       & 5.00\%   & 10.00\%  &       & 5.00\%   & 10.00\% \bigstrut\\
    \hline
 & Ensemble Features \cite{EnsembleFeatures_2020} & 23.66 & 51.45 & 39.96 & 5.82  & 7.22  & 2.92  & 6.17  & 7.54  & 3.94 \bigstrut\\
\cline{2-11}                & Hybrid Features \cite{RagISBA2019} & 47.15 & 87.16 & 79.41 & 6.50   & 8.23  & 4.15  & 7.91  & 10.29 & 6.34 \bigstrut\\

\cline{2-11}       &  Deep Features \cite{raja2017transferable} &  11.63  & 23.11     & 12.56     & 19.91  & 34.10  &  31.38  & 18.66  & 37.75  & 28.64 \bigstrut\\

\cline{2-11}           &  Steerable Features \cite{ramachandra2019detecting} & 50.00  & 91.25     & 85.92   & 35.86  & 80.44  & 70.66 & 38.60  & 84.73  & 74.99 \bigstrut\\

\cline{2-11}       \multicolumn{1}{|c|}{\multirow{-1}{*}{Landmarks-I \cite{raghavendra2017face}}}    & Multi-Modality \cite{raghavendra2022multimodality}  & 32.43  & 73.17      &  61.92      & 9.65   & 15.71  & 8.75   & \textbf{5.01}  & \textbf{5.00}  & \textbf{3.14}\bigstrut\\

\cline{2-11}         & Residual AutoEncoder \cite{raja2022towards}  & 18.70  &  31.62    &  25.43     & 4.12  & 3.95  & 1.89  & 8.02  &  11.51 & 7.39\bigstrut\\

\cline{2-11}         & Multi-level Deep Features \cite{venkatesh2022multi}  & \textbf{6.01}  &  \textbf{6.86}    &  \textbf{3.43}    & \textbf{1.37} & \textbf{0.51}  & \textbf{0.17} &6.01  & 7.37 &  7.37 \bigstrut\\

\cline{2-11}             & \textbf{Proposed Method}  & 7.38  & 11.84     & 5.66     & 8.25  & 14.43  & 6.87  & 12.03  & 19.42  & 13.23\bigstrut\\

\cline{1-11}          & Ensemble Features \cite{EnsembleFeatures_2020} & 35.38 & 82.33 & 68.95 & 41.67 & 95.14 & 83.53 & 43.68 & 96.01 & 85.44 \bigstrut\\
\cline{2-11}                & Hybrid Features \cite{RagISBA2019} & 28.62 & 75.64 & 61.4  & 44.38 & 95.66 & 85.78 & 38.18 & 90.46 & 78.16 \bigstrut\\

\cline{2-11}       &  Deep Features \cite{raja2017transferable} &  38.40  & 89.02     & 77.70    & 45.21  & 86.17  &  80.27  & 42.80  & 96.50  & 90.48 \bigstrut\\

\cline{2-11}           &  Steerable Features \cite{ramachandra2019detecting} & 50.00  & 97.94     & 93.82   & 44.36 & 87.26  & 81.15 & 43.92  & 92.36  & 86.55 \bigstrut\\

\cline{2-11}       \multicolumn{1}{|c|}{\multirow{-1}{*}{Landmarks-II \cite{UBO_Morphing_Tool}}}    & Multi-Modality \cite{raghavendra2022multimodality}  & \textbf{17.14}  & \textbf{49.39}      &  \textbf{29.41}      & \textbf{28.80}   & 87.14  & \textbf{58.92}   & 29.65  & 88.21  & 67.50\bigstrut\\

\cline{2-11}         & Residual AutoEncoder \cite{raja2022towards}  & 57.22  &  90.21    &  85.57     & 32.08  & \textbf{82.82}  & 71.31  &33.51  &  77.66 & 68.73\bigstrut\\

\cline{2-11}         & Multi-level Deep Features \cite{venkatesh2022multi}  & 40.49  &  86.27    &  74.27     & 30.22 & 65.44  & 55.32 &\textbf{6.01}  & \textbf{7.37} &  \textbf{3.43} \bigstrut\\

\cline{2-11}            & \textbf{Proposed Method}  & 32.76  & 83.53     & 73.41     & 39.06  & 89.41  & 79.86  & 35.70  & 87.69  & 76.08\bigstrut\\

\cline{1-11}           & Ensemble Features \cite{EnsembleFeatures_2020} & 17.72 & 37.22 & 26.58 & 12.19 & 26.24 & 15.26 & 11.82 & 24.69 & 14.23 \bigstrut\\
\cline{2-11}                & Hybrid Features \cite{RagISBA2019} & 31.16 & 64.32 & 53.85 & 11.99 & 19.20  & 13.72 & 9.93  & 18.15 & 9.94 \bigstrut\\

\cline{2-11}       &  Deep Features \cite{raja2017transferable} &  26.86  & 27.54     & 24.63     &  \textbf{0.00} & \textbf{0.00}  &  \textbf{0.00}  & 10.46  & 8.74  & 6.43 \bigstrut\\

\cline{2-11}           &  Steerable Features \cite{ramachandra2019detecting} & 50.00  & 92.45     & 87.15   & \textbf{0.00} & \textbf{0.00}  & \textbf{0.00} & 44.12  & 92.45  & 86.79 \bigstrut\\

\cline{2-11}       \multicolumn{1}{|c|}{\multirow{-1}{*}{StyleGAN \cite{MorphStyleGAN2020}}}    & Multi-Modality \cite{raghavendra2022multimodality}  & 22.81  & 48.37      &  37.22      & \textbf{0.00}   & \textbf{0.00}  & \textbf{0.00}   & 5.34  & \textbf{5.53}  & 4.15\bigstrut\\

\cline{2-11}         & Residual AutoEncoder \cite{raja2022towards}  & 10.14  &  15.12    &  10.14     & 11.37  & 19.24  & 12.54  &10.30  & 19.93 & 10.65\bigstrut\\

\cline{2-11}         & Multi-level Deep Features \cite{venkatesh2022multi}  & \textbf{6.33}  &  \textbf{7.54}    &  \textbf{4.11}     & 3.95 & 2.05  & 0.51 &\textbf{5.12}  & 5.66 &  \textbf{2.05} \bigstrut\\

\cline{2-11}            & \textbf{Proposed Method}  & 8.92  & 16.30     & 7.89     &7.89  & 10.98  & 6.69  & 9.09  & 14.41  & 8.58\bigstrut\\

\cline{1-11}          \cellcolor{mygray} & \cellcolor{mygray}Ensemble Features \cite{EnsembleFeatures_2020} & \cellcolor{mygray} \textbf{0.00}     & \cellcolor{mygray} \textbf{0.00}     & \cellcolor{mygray} \textbf{0.00}     & \cellcolor{mygray} \textbf{0.00}     & \cellcolor{mygray} \textbf{0.00}     & \cellcolor{mygray} \textbf{0.00}     & \cellcolor{mygray} \textbf{0.00}     & \cellcolor{mygray} \textbf{0.00}     & \cellcolor{mygray} \textbf{0.00} \bigstrut\\
\cline{2-11}          \cellcolor{mygray}      & \cellcolor{mygray}Hybrid Features \cite{RagISBA2019} & \cellcolor{mygray} \textbf{0.00}     & \cellcolor{mygray} \textbf{0.00}     & \cellcolor{mygray} \textbf{0.00}     & \cellcolor{mygray} \textbf{0.00}     & \cellcolor{mygray} \textbf{0.00}     & \cellcolor{mygray} \textbf{0.00}     & \cellcolor{mygray} \textbf{0.00}     & \cellcolor{mygray} \textbf{0.00}     & \cellcolor{mygray} \textbf{0.00} \bigstrut\\

\cline{2-11}         \cellcolor{mygray}   &  \cellcolor{mygray}Deep Features \cite{raja2017transferable} & \cellcolor{mygray} 1.16  & \cellcolor{mygray} \textbf{0.00}     & \cellcolor{mygray} \textbf{0.00}     & \cellcolor{mygray} \textbf{0.00}  & \cellcolor{mygray} \textbf{0.00}  & \cellcolor{mygray} \textbf{0.00}  & \cellcolor{mygray}\textbf{0.00}  & \cellcolor{mygray}\textbf{0.00}  & \cellcolor{mygray} \textbf{0.00} \bigstrut\\

\cline{2-11}         \cellcolor{mygray}   &  \cellcolor{mygray}Steerable Features \cite{ramachandra2019detecting} & \cellcolor{mygray} 29.83  & \cellcolor{mygray} 84.21     & \cellcolor{mygray} 68.95   & \cellcolor{mygray} \textbf{0.00}  & \cellcolor{mygray} \textbf{0.00}  & \cellcolor{mygray} \textbf{0.00} & \cellcolor{mygray}25.92  & \cellcolor{mygray}72.72  & \cellcolor{mygray} 60.14 \bigstrut\\

\cline{2-11}      \multicolumn{1}{|c|}{\multirow{-1}{*}{\cellcolor{mygray}MIPGAN-I \cite{zhang2021mipgan}}}    & \cellcolor{mygray}Multi-Modality \cite{raghavendra2022multimodality}  & \cellcolor{mygray}\textbf{0.00}  & \cellcolor{mygray} \textbf{0.00}     & \cellcolor{mygray} \textbf{0.00}     & \cellcolor{mygray}\textbf{0.00}  & \cellcolor{mygray}\textbf{0.00}  & \cellcolor{mygray} \textbf{0.00}  & \cellcolor{mygray}\textbf{0.00}  & \cellcolor{mygray} \textbf{0.00} & \cellcolor{mygray} \textbf{0.00} \bigstrut\\

\cline{2-11}      \cellcolor{mygray}    & \cellcolor{mygray}Residual AutoEncoder \cite{raja2022towards}  & \cellcolor{mygray}1.03  & \cellcolor{mygray} \textbf{0.00}     & \cellcolor{mygray} \textbf{0.00}     & \cellcolor{mygray}0.34  & \cellcolor{mygray}\textbf{0.00}  & \cellcolor{mygray}\textbf{0.00}  & \cellcolor{mygray} 0.69  & \cellcolor{mygray} \textbf{0.00}  & \cellcolor{mygray} \textbf{0.00} \bigstrut\\

\cline{2-11}      \cellcolor{mygray}    & \cellcolor{mygray} Multi-level Deep Features \cite{venkatesh2022multi}  & \cellcolor{mygray}1.36  & \cellcolor{mygray} 0.17     & \cellcolor{mygray} \textbf{0.00}     & \cellcolor{mygray}11.35  & \cellcolor{mygray}29.84  & \cellcolor{mygray}14.57  & \cellcolor{mygray} 1.04  & \cellcolor{mygray} 1.17  & \cellcolor{mygray} 0.17 \bigstrut\\

\cline{2-11}         \cellcolor{mygray}    & \cellcolor{mygray}\textbf{Proposed Method}  & \cellcolor{mygray} 0.51  & \cellcolor{mygray} \textbf{0.00}     & \cellcolor{mygray} \textbf{0.00}     & \cellcolor{mygray}0.86  & \cellcolor{mygray}0.17  & \cellcolor{mygray}0.17  & \cellcolor{mygray}1.20  & \cellcolor{mygray}0.17  & \cellcolor{mygray}0.17\bigstrut\\
\cline{1-11}           & Ensemble Features \cite{EnsembleFeatures_2020} & 2.15  & 0.17  & \textbf{0.00}     & 0.68  & \textbf{0.00}     & \textbf{0.00}     & 0.64  & \textbf{0.00}     & \textbf{0.00} \bigstrut\\
\cline{2-11}               & Hybrid Features \cite{RagISBA2019} & 1.36  & 0.34  & \textbf{0.00}     & 0.86  & \textbf{0.00}     & \textbf{0.00}     & 0.85 & \textbf{0.00}     & \textbf{0.00} \bigstrut\\

\cline{2-11}       &  Deep Features \cite{raja2017transferable} &  2.01  & 1.02     & 0.34     & \textbf{0.00}  & \textbf{0.00}  &  \textbf{0.00}  & \textbf{0.00}  & \textbf{0.00}  & \textbf{0.00} \bigstrut\\

\cline{2-11}           &  Steerable Features \cite{ramachandra2019detecting} & 34.56  & 88.16     & 78.38   & \textbf{0.00}  & \textbf{0.00}  & \textbf{0.00} & 28.58  & 85.93  & 69.12 \bigstrut\\

\cline{2-11}       \multicolumn{1}{|c|}{\multirow{-1}{*}{MIPGAN-II \cite{zhang2021mipgan}}}    & Multi-Modality \cite{raghavendra2022multimodality}  & \textbf{0.00}  & \textbf{0.00}      &  \textbf{0.00}      & \textbf{0.00}   & \textbf{0.00}  & \textbf{0.00}   & \textbf{0.00}  & \textbf{0.00}  & \textbf{0.00}\bigstrut\\

\cline{2-11}         & Residual AutoEncoder \cite{raja2022towards}  & 0.03  &  \textbf{0.00}    &  \textbf{0.00}     & 0.30  & \textbf{0.00}  & \textbf{0.00}  &0.53 &  \textbf{0.00} & \textbf{0.00}\bigstrut\\

\cline{2-11}         & Multi-level Deep Features \cite{venkatesh2022multi}  & 1.04  &  \textbf{0.00}    &  \textbf{0.00}     & 3.06& 0.51  & 0.17&0.48  & 0.34 &  0.17 \bigstrut\\

\cline{2-11}         & \textbf{Proposed Method}  & 2.57  & 0.69     & 0.34     & 0.34  & 0.17  & \textbf{0.00}  & 0.51  & \textbf{0.00}  & \textbf{0.00}\bigstrut\\

    \hline
    \end{tabular}%
    }
  \label{tab:MADMIPGAN1}%
\end{table*}%


\begin{table*}[h!]
  \centering
  \caption{Quantitative performance of MAD on FRGC morph database. Training morphing type: MIPGAN-II \cite{zhang2021mipgan}.}
  \resizebox{1\linewidth}{!}{
    \begin{tabular}{|c|c|c|c|c|c|c|c|c|c|c|}
    \hline
  \textbf{Testing}& \multirow{3}[6]{*}{\textbf{MAD Algorithms}} & \multicolumn{3}{c|}{\textbf{Digital}} & \multicolumn{3}{c|}{\textbf{Print-scan}} & \multicolumn{3}{c|}{\textbf{Print-scan with compression}} \bigstrut\\
\cline{3-11}           \textbf{Morphing Type}       & \multicolumn{1}{c|}{} & \multicolumn{1}{c|}{\multirow{2}[4]{*}{D-EER(\%)}} & \multicolumn{2}{c|}{BPCER @ MACER =  } & \multicolumn{1}{c|}{\multirow{2}[4]{*}{D-EER(\%)}} & \multicolumn{2}{c|}{BPCER @ MACER =  } & \multicolumn{1}{c|}{\multirow{2}[4]{*}{D-EER(\%)}} & \multicolumn{2}{c|}{BPCER @ MACER =  } \bigstrut\\
\cline{4-5}\cline{7-8}\cline{10-11}      \multirow{3}[6]{*}{}          & \multicolumn{1}{c|}{} &       & 5.00\%   & 10.00\%  &       & 5.00\%   & 10.00\%  &       & 5.00\%   & 10.00\% \bigstrut\\
    \hline
  & Ensemble Features \cite{EnsembleFeatures_2020} & 13.08 & 29.15 & 15.78 & \textbf{4.28}  & \textbf{3.94}  & \textbf{2.22}  & 4.28  & 3.61  & 2.22 \bigstrut\\
\cline{2-11}                 & Hybrid Features \cite{RagISBA2019} & 40.14 & 77.70  & 67.23 & 5.49  & 5.48  & 2.40   & 7.21  & 10.98 & 4.15 \bigstrut\\

\cline{2-11}           &  Deep Features \cite{raja2017transferable} &  16.31  & 39.96     &  26.17     & 18.66  &  37.73  &  28.64  & 8.17  & 13.89  &  6.88 \bigstrut\\

\cline{2-11}            &  Steerable Features \cite{ramachandra2019detecting} &  50.00  & 91.25     &  86.27   & 35.86  &  80.44  &  70.66 & 39.09  & 89.87  & 79.75 \bigstrut\\

\cline{2-11}       \multicolumn{1}{|c|}{\multirow{-1}{*}{Landmarks-I \cite{raghavendra2017face}}}    & Multi-Modality \cite{raghavendra2022multimodality}  & 23.82  & 54.20     &  37.90      & 6.18  & 8.21  & 3.21   & \textbf{4.16}  & \textbf{3.39}  & \textbf{1.96}\bigstrut\\

\cline{2-11}         & Residual AutoEncoder \cite{raja2022towards}  & 50.00  &  82.47    &  74.57    & 11.22  & 17.53  & 12.37  & 20.27  & 38.32 & 30.58\bigstrut\\

\cline{2-11}         & Multi-level Deep Features \cite{venkatesh2022multi}  & \textbf{6.21}  &  \textbf{6.51}    &  \textbf{4.80}     & 14.25& 30.70 & 18.86 &6.70  & 8.06 &  3.94 \bigstrut\\

\cline{2-11}            & \textbf{Proposed Method}  & 7.20 & 10.12     & 5.66     & 11.51  & 24.05  & 14.43  & 36.77  & 1.12  & 21.13\bigstrut\\

\cline{1-11}           & Ensemble Features \cite{EnsembleFeatures_2020} & 32.37 & 84.90  & 70.32 & 39.20  & 90.12 & 82.32 & 44.17 & 95.49 & 88.73 \bigstrut\\
\cline{2-11}                 & Hybrid Features \cite{RagISBA2019} & 23.88 & 63.80  & 45.62 & 40.22 & 88.90  & 79.20  & 38.96 & 94.28 & 82.14 \bigstrut\\

\cline{2-11}       &  Deep Features \cite{raja2017transferable} &  41.10  & 90.92     & 83.87     & 42.81  & 96.50  & 90.40  & 35.12  & \textbf{76.56}  & 70.13 \bigstrut\\

\cline{2-11}           &  Steerable Features \cite{ramachandra2019detecting} & 48.94  & 97.77     & 92.79   & 44.36  & 87.26  & 81.15 & 45.47  & 92.53  & 88.71 \bigstrut\\

\cline{2-11}       \multicolumn{1}{|c|}{\multirow{-1}{*}{Landmarks-II \cite{UBO_Morphing_Tool}}}    & Multi-Modality \cite{raghavendra2022multimodality}  & \textbf{10.62}  & \textbf{28.81}      &  \textbf{11.32}    & \textbf{26.80}   & 82.32  & \textbf{64.28}   & \textbf{30.15}  & 90.53  & \textbf{73.21}\bigstrut\\

\cline{2-11}         & Residual AutoEncoder \cite{raja2022towards}  & 64.75  & 82.47    & 82.47  & 45.88  & \textbf{79.38}  & 79.38  & 51.03  &  80.07 & 80.07\bigstrut\\

\cline{2-11}         & Multi-level Deep Features \cite{venkatesh2022multi}  & 46.13  &  95.88    &  99.90     & 42.03 & 97.90  & 93.19 &39.24  & 84.20 &  74.13 \bigstrut\\

\cline{2-11}             & \textbf{Proposed Method}  & 35.33  & 83.36    & 74.27     & 44.79  & 92.36  & 87.85  & 42.11  & 90.99  & 83.36\bigstrut\\

\cline{1-11}           & Ensemble Features \cite{EnsembleFeatures_2020} & 12.51 & 22.29 & 15.78 & 13.72 & 29.67 & 18.18 & 14.25 & 31.73 & 20.41 \bigstrut\\
\cline{2-11}                 & Hybrid Features \cite{RagISBA2019} & 24.70  & 49.74 & 41.85 & 12.87 & 26.58 & 14.75 & 11.86 & 26.92 & 15.09 \bigstrut\\

\cline{2-11}       &  Deep Features \cite{raja2017transferable} &  21.70  & 33.49     & 23.72     & \textbf{0.00}  & \textbf{0.00}  &  \textbf{0.00}  & 7.18  & 9.43  & 4.45 \bigstrut\\

\cline{2-11}           &  Steerable Features \cite{ramachandra2019detecting} & 50.00  & 95.19     & 92.45   & \textbf{0.00}  & \textbf{0.00}  & \textbf{0.00} & 46.30  & 91.54  & 90.56 \bigstrut\\

\cline{2-11}       \multicolumn{1}{|c|}{\multirow{-1}{*}{StyleGAN \cite{MorphStyleGAN2020}}}    & Multi-Modality \cite{raghavendra2022multimodality}  & 21.15  & 40.48      &  30.87      &  \textbf{0.00}   &  \textbf{0.00}  &  \textbf{0.00}   & 6.79  & 9.28  & 3.92\bigstrut\\

\cline{2-11}         & Residual AutoEncoder \cite{raja2022towards}  & 33.33  &  72.34    &  64.95     & 13.81  & 25.26  & 17.01  & 11.68  &  20.45 & 13.75\bigstrut\\

\cline{2-11}         & Multi-level Deep Features \cite{venkatesh2022multi}  & \textbf{5.52}  &  \textbf{6.51}    &  \textbf{3.25}     & \textbf{0.00} & \textbf{0.00}  & \textbf{0.00} &\textbf{6.01}  & \textbf{7.03}& \textbf{3.25} \bigstrut\\

\cline{2-11}             & \textbf{Proposed Method}  & 14.07  & 23.67     & 16.64     & 9.78  & 17.15  & 9.09  & 10.63  & 24.53  & 11.66\bigstrut\\

\cline{1-11}           & Ensemble Features \cite{EnsembleFeatures_2020} & 1.56  & 0.68  & 0.34  & 2.14  & 1.22  & 0.53  & 2.57  & 0.85  & 0.34 \bigstrut\\
\cline{2-11}                 & Hybrid Features \cite{RagISBA2019} & 2.27  & 0.85  & 0.17  & 4.79  & 4.80   & 3.43  & 4.30   & 3.60   & 2.22 \bigstrut\\

\cline{2-11}       &  Deep Features \cite{raja2017transferable} &  2.41  & 0.55    & 0.17     & \textbf{0.00}  & \textbf{0.00}  &  \textbf{0.00}  & 7.85  & 10.97  & 5.31 \bigstrut\\

\cline{2-11}           &  Steerable Features \cite{ramachandra2019detecting} & 30.51  & 83.70     & 69.29   & \textbf{0.00}  & \textbf{0.00}  & \textbf{0.00} & 26.24  & 82.33  & 62.63 \bigstrut\\

\cline{2-11}       \multicolumn{1}{|c|}{\multirow{-1}{*}{MIPGAN-I \cite{zhang2021mipgan}}}    & Multi-Modality \cite{raghavendra2022multimodality}  & \textbf{0.00}  & \textbf{0.00}      &  \textbf{0.00}      & \textbf{0.00}   & \textbf{0.00}  & \textbf{0.00}   & \textbf{0.00}  & \textbf{0.00}  & \textbf{0.00}\bigstrut\\

\cline{2-11}         & Residual AutoEncoder \cite{raja2022towards}  & 1.37  &  0.68    &  0.17     & 45.95  & 79.38  & 79.38  & 35.51  & 80.07 & 72.51\bigstrut\\

\cline{2-11}         & Multi-level Deep Features \cite{venkatesh2022multi}  & 2.05  &  0.85    &  0.34     & \textbf{0.00} & \textbf{0.00}  & \textbf{0.00} &4.30  & 3.60& 2.05 \bigstrut\\

\cline{2-11}             & \textbf{Proposed Method}  & 0.86  & 0.17     & \textbf{0.00}     & 17.15  & 39.45  & 26.59  & 13.55  & 26.59  & 16.64\bigstrut\\

\cline{1-11}           \cellcolor{mygray} & \cellcolor{mygray}Ensemble Features \cite{EnsembleFeatures_2020} & \cellcolor{mygray} \textbf{0.00}     & \cellcolor{mygray} \textbf{0.00}     & \cellcolor{mygray} \textbf{0.00}     & \cellcolor{mygray} \textbf{0.00}     & \cellcolor{mygray} \textbf{0.00}     & \cellcolor{mygray} \textbf{0.00}     & \cellcolor{mygray} \textbf{0.00}     & \cellcolor{mygray} \textbf{0.00}     & \cellcolor{mygray} \textbf{0.00} \bigstrut\\
\cline{2-11}       \cellcolor{mygray}         & \cellcolor{mygray}Hybrid Features \cite{RagISBA2019} & \cellcolor{mygray} \textbf{0.00}     & \cellcolor{mygray} \textbf{0.00}     & \cellcolor{mygray} \textbf{0.00}     & \cellcolor{mygray} \textbf{0.00}     & \cellcolor{mygray} \textbf{0.00}     & \cellcolor{mygray} \textbf{0.00}     & \cellcolor{mygray} \textbf{0.00}     & \cellcolor{mygray} \textbf{0.00}     & \cellcolor{mygray} \textbf{0.00} \bigstrut\\

\cline{2-11}         \cellcolor{mygray}   &  \cellcolor{mygray}Deep Features \cite{raja2017transferable} & \cellcolor{mygray} 2.57  & \cellcolor{mygray} 1.02     & \cellcolor{mygray} 0.51     & \cellcolor{mygray} \textbf{0.00} & \cellcolor{mygray}\textbf{0.00}  & \cellcolor{mygray} \textbf{0.00}  & \cellcolor{mygray}3.58  & \cellcolor{mygray}1.20  & \cellcolor{mygray} \textbf{0.00} \bigstrut\\

\cline{2-11}         \cellcolor{mygray}   &  \cellcolor{mygray}Steerable Features \cite{ramachandra2019detecting} & \cellcolor{mygray} 31.84  & \cellcolor{mygray} 86.79     & \cellcolor{mygray} 70.84   & \cellcolor{mygray} \textbf{0.00}  & \cellcolor{mygray}\textbf{0.00}  & \cellcolor{mygray} \textbf{0.00} & \cellcolor{mygray}26.57  & \cellcolor{mygray}85.59  & \cellcolor{mygray} 72.21 \bigstrut\\

\cline{2-11}      \multicolumn{1}{|c|}{\multirow{-1}{*}{\cellcolor{mygray}MIPGAN-II \cite{zhang2021mipgan}}}    & \cellcolor{mygray}Multi-Modality \cite{raghavendra2022multimodality}  & \cellcolor{mygray}\textbf{0.00}  & \cellcolor{mygray} \textbf{0.00}     & \cellcolor{mygray} \textbf{0.00}     & \cellcolor{mygray}\textbf{0.00}  & \cellcolor{mygray}\textbf{0.00}  & \cellcolor{mygray} \textbf{0.00}  & \cellcolor{mygray} \textbf{0.00}  & \cellcolor{mygray} \textbf{0.00}  & \cellcolor{mygray} \textbf{0.00} \bigstrut\\

\cline{2-11}      \cellcolor{mygray}    & \cellcolor{mygray}Residual AutoEncoder \cite{raja2022towards}  & \cellcolor{mygray} 0.86  & \cellcolor{mygray} 0.17     & \cellcolor{mygray} 0.17     & \cellcolor{mygray} 8.73  & \cellcolor{mygray} 29.90  & \cellcolor{mygray} 2.58  & \cellcolor{mygray} 8.73  & \cellcolor{mygray} 31.79  & \cellcolor{mygray} 2.92 \bigstrut\\

\cline{2-11}      \cellcolor{mygray}    & \cellcolor{mygray}Multi-level Deep Features \cite{venkatesh2022multi}  & \cellcolor{mygray}1.85  & \cellcolor{mygray} 0.34     & \cellcolor{mygray} \textbf{0.00}     & \cellcolor{mygray}\textbf{0.00}  & \cellcolor{mygray}\textbf{0.00}  & \cellcolor{mygray}\textbf{0.00}  & \cellcolor{mygray} 2.53  & \cellcolor{mygray} 0.85  & \cellcolor{mygray} 3.25 \bigstrut\\

\cline{2-11}     \cellcolor{mygray}    & \cellcolor{mygray}\textbf{Proposed Method}  & \cellcolor{mygray}0.69  & \cellcolor{mygray}\textbf{0.00}     & \cellcolor{mygray} \textbf{0.00}     & \cellcolor{mygray}5.49  & \cellcolor{mygray}6.35  & \cellcolor{mygray}0.17  & \cellcolor{mygray}4.46  & \cellcolor{mygray}3.26  & \cellcolor{mygray}0.17\bigstrut\\
    \hline
    \end{tabular}%
    }
  \label{tab:MADMIPGAN2}%
\end{table*}%

\subsection{Separated Analysis on Intra and Inter Dataset Testing}
In this section, we split the cases of intra-dataset (training and testing with morphs generated by the same algorithm) and inter-dataset testing (training and testing with morphs generated by the different algorithms). Then we apply the statistical analysis similar to the process \cref{sec:exp}. As shown in \cref{tab:statistical_inter}, the proposed method shows state-of-the-art detection accuracy on digital images. Regarding to the intra-dataset testing results reported in \cref{tab:statistical_intra}, it is also shown that the proposed method has comparable performance benchmarking with other baselines.

\begin{table*}[hbtp]
\centering
 \caption{Statistical analysis on the D-EER($\%$) computed for all inter-dataset testing results on FRGC morphing dataset.}
\resizebox{0.75\linewidth}{!}{
\begin{tabular}{|c|p{1.1cm}<{\centering}|p{1cm}<{\centering}|p{1cm}<{\centering}|p{1cm}<{\centering}|p{1.4cm}<{\centering}|p{1.4cm}<{\centering}|}
\hline
\textbf{S-MAD} & \multicolumn{2}{c|}{\textbf{Digital}}    & \multicolumn{2}{c|}{\textbf{Print-scan}} & \multicolumn{2}{c|}{\textbf{P.S. with Compression}} \\ 
\cline{2-7} \textbf{Algorithms} & $\mu$ & $\sigma$ & $\mu$ & $\sigma$ & $\mu$    & $\sigma$   \\ \hline
Ensemble & \multicolumn{1}{c|}{26.10} & 16.00 & \multicolumn{1}{c|}{20.48} & 14.35 & \multicolumn{1}{c|}{21.14} & 15.29 \\ 
Features \cite{EnsembleFeatures_2020} &  & &  &  &  &  \\ \hline
Hybrid   & \multicolumn{1}{c|}{32.56} & 17.48 & \multicolumn{1}{c|}{23.45} & 14.94 & \multicolumn{1}{c|}{19.97} & 13.56 \\ 
Features \cite{RagISBA2019}  &  &  &  &  &  &  \\ \hline
Deep    & \multicolumn{1}{c|}{27.51} & 16.21 & \multicolumn{1}{c|}{13.75} & 17.31 & \multicolumn{1}{c|}{20.32} & 12.35 \\ 
 Features \cite{raja2017transferable}  &  &  &  &  &  &  \\ \hline
 Steerable    & \multicolumn{1}{c|}{43.25} & 14.18 & \multicolumn{1}{c|}{17.17} & 19.28 & \multicolumn{1}{c|}{36.78} & \textbf{6.16} \\ 
 Features\cite{ramachandra2019detecting}  &  &  &  &  &  &  \\ \hline
Multi-   & \multicolumn{1}{c|}{22.57} & 14.11 & \multicolumn{1}{c|}{\textbf{9.46}} & 13.10 & \multicolumn{1}{c|}{16.61} & 13.73 \\ 
Modality \cite{raghavendra2022multimodality}  &  &  &  &  &  &  \\ \hline
Residual   & \multicolumn{1}{c|}{20.81} & 19.63 & \multicolumn{1}{c|}{20.22} & 18.15 & \multicolumn{1}{c|}{17.66} & 14.22 \\ 
AutoEncoder \cite{raja2022towards}  &  &  &  &  &  &  \\ \hline
Multi-level & \multicolumn{1}{c|}{17.79} & 13.52 & \multicolumn{1}{c|}{11.51} & \textbf{12.27}& \multicolumn{1}{c|}{\textbf{15.89}} & 12.27  \\ 
Deep Feature \cite{venkatesh2022multi}  &  &  &  &  &  &  \\ \hline
\textbf{Proposed Method}   & \multicolumn{1}{c|}{\textbf{16.41}} & \textbf{11.20} & \multicolumn{1}{c|}{21.86} & 13.78 & \multicolumn{1}{c|}{22.85} & 12.49 \\ \hline
\end{tabular}
}
\label{tab:statistical_inter}%
\end{table*}

\begin{table*}[hbtp]
\centering
 \caption{Statistical analysis on the D-EER($\%$) computed for all intra-dataset testing results on FRGC morphing dataset.}
\resizebox{0.75\linewidth}{!}{
\begin{tabular}{|c|p{1.1cm}<{\centering}|p{1cm}<{\centering}|p{1cm}<{\centering}|p{1cm}<{\centering}|p{1.4cm}<{\centering}|p{1.4cm}<{\centering}|}
\hline
\textbf{S-MAD} & \multicolumn{2}{c|}{\textbf{Digital}}    & \multicolumn{2}{c|}{\textbf{Print-scan}} & \multicolumn{2}{c|}{\textbf{P.S. with Compression}} \\ 
\cline{2-7} \textbf{Algorithms} & $\mu$ & $\sigma$ & $\mu$ & $\sigma$ & $\mu$    & $\sigma$   \\ \hline
Ensemble & \multicolumn{1}{c|}{0.72} & 1.45 & \multicolumn{1}{c|}{1.73} & 2.47 & \multicolumn{1}{c|}{1.63} & 2.21 \\ 
Features \cite{EnsembleFeatures_2020} &  & &  &  &  &  \\ \hline
Hybrid   & \multicolumn{1}{c|}{0.34} & \textbf{0.60} & \multicolumn{1}{c|}{1.41} & 2.03 & \multicolumn{1}{c|}{1.52} & 2.11 \\ 
Features \cite{RagISBA2019}  &  &  &  &  &  &  \\ \hline
Deep    & \multicolumn{1}{c|}{1.98} & 2.29 & \multicolumn{1}{c|}{1.90} & 2.62 & \multicolumn{1}{c|}{2.49} & 2.63 \\ 
 Features \cite{raja2017transferable}  &  &  &  &  &  &  \\ \hline
 Steerable    & \multicolumn{1}{c|}{19.74} & 12.42 & \multicolumn{1}{c|}{10.83} & 13.44 & \multicolumn{1}{c|}{25.97} & 2.92 \\ 
 Features\cite{ramachandra2019detecting}  &  &  &  &  &  &  \\ \hline
Multi-   & \multicolumn{1}{c|}{\textbf{0.32}} & 0.64 & \multicolumn{1}{c|}{\textbf{0.32}} & \textbf{0.64} & \multicolumn{1}{c|}{\textbf{0.65}} & \textbf{0.80} \\ 
Modality \cite{raghavendra2022multimodality}  &  &  &  &  &  &  \\ \hline
Residual   & \multicolumn{1}{c|}{0.89} & 0.94 & \multicolumn{1}{c|}{2.03} & 3.00 & \multicolumn{1}{c|}{3.97} & 4.16 \\ 
AutoEncoder \cite{raja2022towards}  &  &  &  &  &  &  \\ \hline
Multi-level & \multicolumn{1}{c|}{2.77} & 4.00 & \multicolumn{1}{c|}{4.41} & 3.95& \multicolumn{1}{c|}{3.32} & 2.06  \\ 
Deep Feature \cite{venkatesh2022multi}  &  &  &  &  &  &  \\ \hline
\textbf{Proposed Method}   & \multicolumn{1}{c|}{2.47} & 4.09 & \multicolumn{1}{c|}{4.21} & 3.99 & \multicolumn{1}{c|}{3.93} & 3.69 \\ \hline
\end{tabular}
}
\label{tab:statistical_intra}%
\end{table*}


\end{document}

%% file: preamble.tex
\usepackage[dvipsnames]{xcolor}
